\documentclass[10pt,letterpaper]{article}

\usepackage[nonatbib,final]{neurips_2020}
\usepackage{times}
\usepackage{epsfig}
\usepackage{graphicx}
\usepackage{amsmath}
\usepackage{amssymb}
\usepackage{multirow}
\usepackage{multicol}
\usepackage{verbatim}
\usepackage{color}
\usepackage[table]{xcolor}
\usepackage{bbm}
\usepackage{booktabs}
\usepackage{enumitem}
\usepackage[font=small]{caption}
\usepackage{subcaption}
\usepackage{wrapfig}

\definecolor{blue}{rgb}{1,0,0}

\usepackage[pagebackref=true,breaklinks=true,letterpaper=true,colorlinks,bookmarks=false]{hyperref}

\makeatletter
\newcommand*\bigcdot{\mathpalette\bigcdot@{.5}}
\newcommand*\bigcdot@[2]{\mathbin{\vcenter{\hbox{\scalebox{#2}{$\m@th#1\bullet$}}}}}
\makeatother

\newcommand{\later}[1]{}
\newcommand{\longer}[1]{}	%

\def\BE{\begin{equation}}
\def\EE{\end{equation}}
\def\BEA{\begin{eqnarray}}
\def\EEA{\end{eqnarray}}
\def\BEAS{\begin{eqnarray*}}
\def\EEAS{\end{eqnarray*}}

\begin{document}

\title{Supervised Contrastive Learning}

{ 
\author{\small Prannay Khosla \thanks{\scriptsize Equal contribution.} \\
{\small Google Research}
\and
{\small \textbf{Piotr Teterwak} \footnotemark[1]\ \  \thanks{\scriptsize Work done while at Google Research.}}\\
{\small Boston University} \\
\and
{\small \textbf{Chen Wang} \footnotemark[2]} \\
{\small Snap Inc.} \\
\and
{\small \textbf{Aaron Sarna} \thanks{\scriptsize Corresponding author: sarna@google.com}} \\
{\small Google Research} \\
\and
{\small \textbf{Yonglong Tian}} \footnotemark[2]\\
{\small MIT} \\
\and
{\small \textbf{Phillip Isola}} \footnotemark[2]\\ 
{\small MIT} \\
\and
{\small \textbf{Aaron Maschinot}} \\
{\small Google Research}\\
\and
{\small \textbf{Ce Liu}} \\
{\small Google Research} \\
\and
{\small \textbf{Dilip Krishnan}} \\
{\small Google Research}\\
}
}


\maketitle
\begin{abstract}
Contrastive learning applied to self-supervised representation learning has seen a resurgence in recent years, leading to state of the art performance in the unsupervised training of deep image models. Modern batch contrastive approaches subsume or significantly outperform traditional contrastive losses such as triplet, max-margin and the N-pairs loss. In this work, we extend the self-supervised batch contrastive approach to the \emph{fully-supervised} setting, allowing us to effectively leverage label information. Clusters of points belonging to the same class are pulled together in embedding space, while simultaneously pushing apart clusters of samples from different classes. We analyze two possible versions of the supervised contrastive (SupCon) loss, identifying the best-performing formulation of the loss. On ResNet-200, we achieve top-1 accuracy of $81.4\%$ on the ImageNet dataset, which is $0.8\%$ above the best number reported for this architecture. We show consistent outperformance over cross-entropy on other datasets and two ResNet variants. The loss shows benefits for robustness to natural corruptions, and is more stable to hyperparameter settings such as optimizers and data augmentations. 
Our loss function is simple to implement and reference TensorFlow code is released at \url{https://t.ly/supcon} \footnote[1]{\scriptsize PyTorch implementation: \url{https://github.com/HobbitLong/SupContrast}}.

\end{abstract}

\section{Introduction}
\begin{wrapfigure}{r}{0.45\textwidth}
\vspace{-35pt}
\centering
\includegraphics[width=\linewidth]{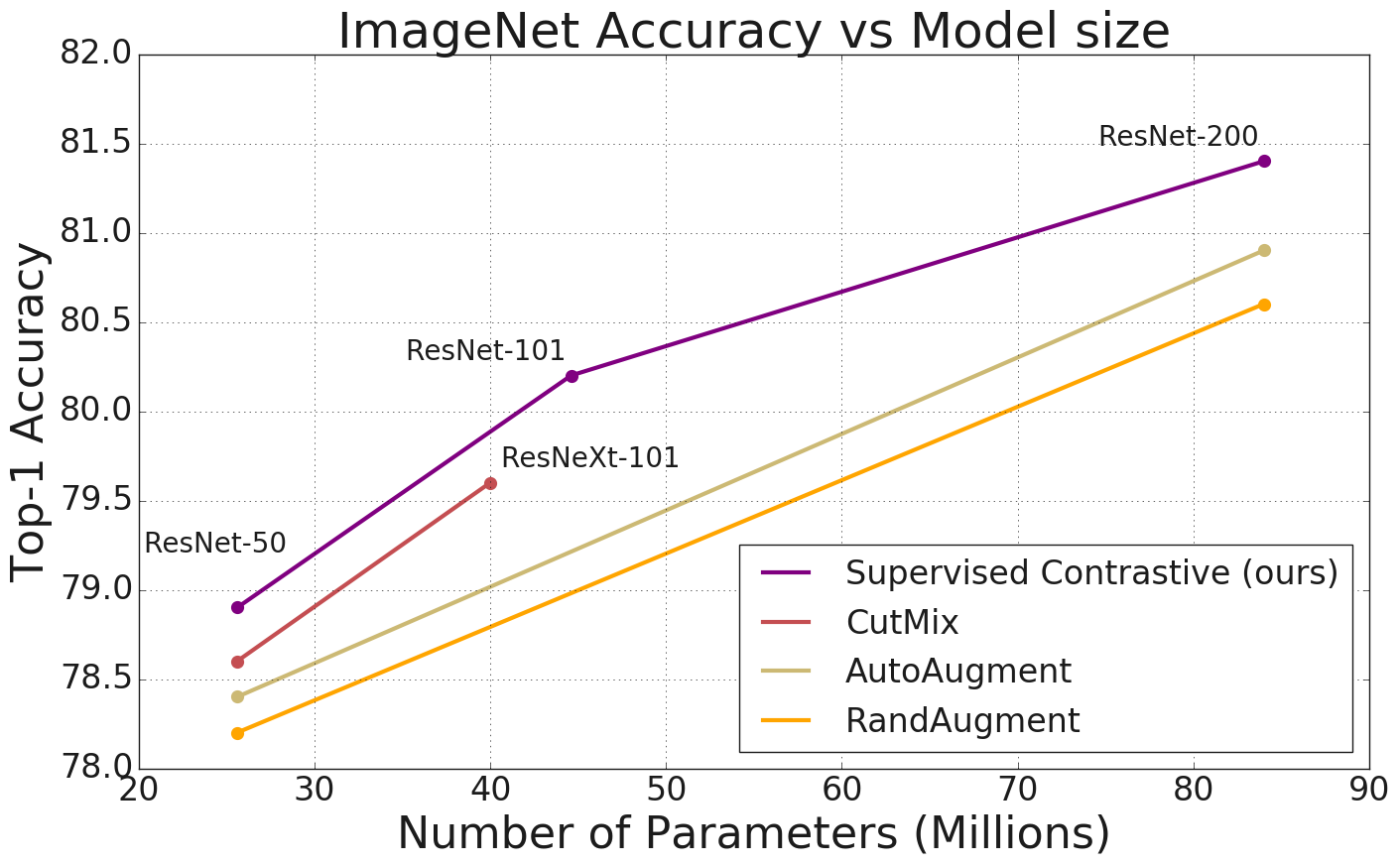}
{\caption{\small Our SupCon loss consistently outperforms cross-entropy with standard data augmentations. We show top-1 accuracy for the ImageNet dataset, on ResNet-50, ResNet-101 and ResNet-200, and compare against AutoAugment \cite{cubuk2019autoaugment}, RandAugment \cite{cubuk2019randaugment} and CutMix \cite{yun2019cutmix}. 
}
\label{fig:imagenet_top1_teaser} }
\vspace{-47pt}
\end{wrapfigure}

\begin{figure*}[t!]  
 \includegraphics[width=\linewidth]{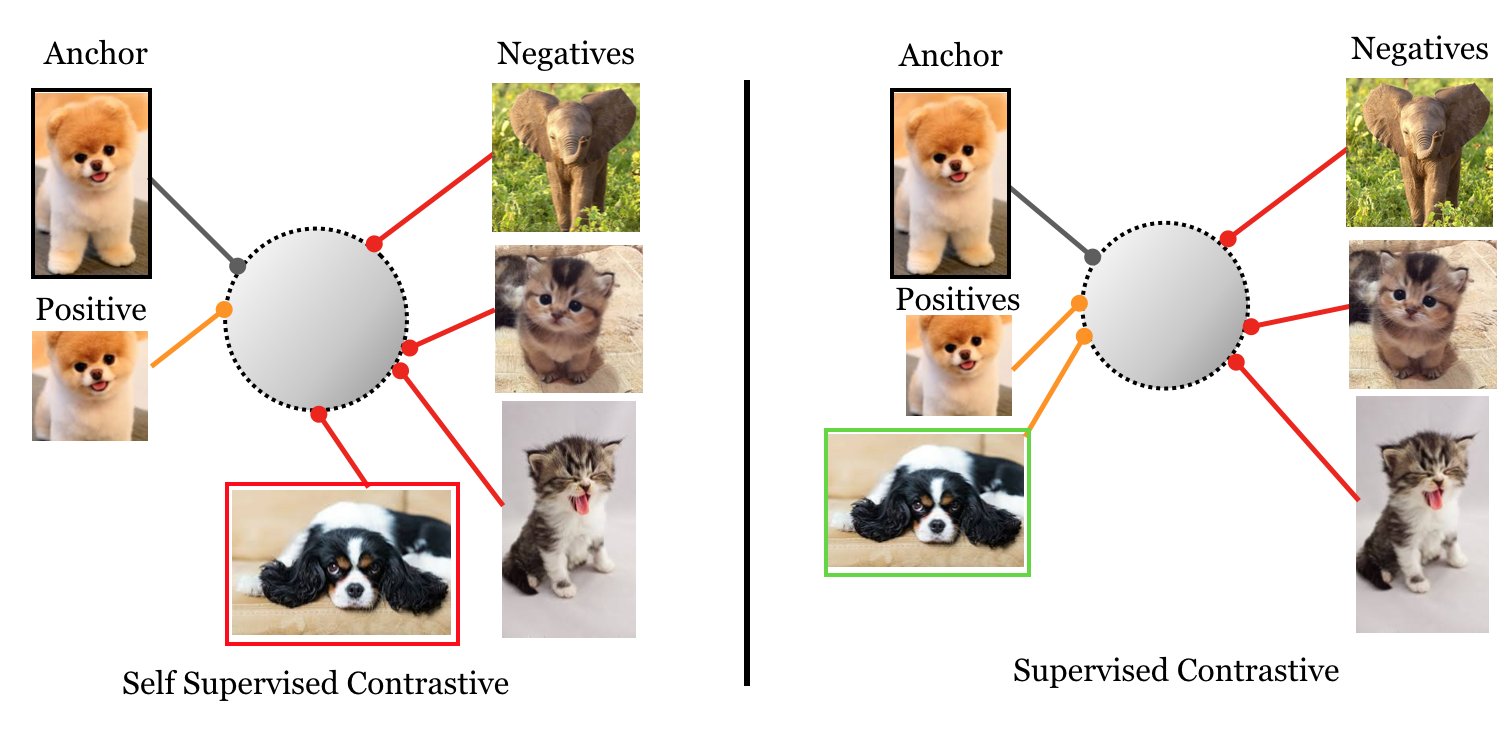}
 {\caption{\small Supervised vs. self-supervised contrastive losses: The self-supervised contrastive loss (left, Eq. \ref{eqn:self_loss}) contrasts a \emph{single} positive for each anchor (i.e., an augmented version of the same image) against a set of negatives consisting of the entire remainder of the batch. The supervised contrastive loss (right) considered in this paper (Eq. \ref{eqn:supervised_loss}), however, contrasts the set of \emph{all} samples from the same class as positives against the negatives from the remainder of the batch. As demonstrated by the photo of the black and white puppy, taking class label information into account results in an embedding space where elements of the same class are more closely aligned than in the self-supervised case.}  
 \label{fig:teaser1}
}
 \vspace{-20pt}
\end{figure*}

The cross-entropy loss is the most widely used loss function for supervised learning of deep classification models. A number of works have explored shortcomings of this loss, such as lack of robustness to noisy labels \cite{zhang2018generalized,sukhbaatar2014training} and the possibility of poor margins \cite{elsayed2018large,liu2016large}, leading to reduced generalization performance. However, in practice, most proposed alternatives have not worked better for large-scale datasets, such as ImageNet \cite{deng2009imagenet}, as evidenced by the continued use of cross-entropy to achieve state of the art results \cite{cubuk2019autoaugment,cubuk2019randaugment,xie2019self,kolesnikov2019large}.

In recent years, a resurgence of work in contrastive learning has led to major advances in self-supervised representation learning \cite{wu2018unsupervised,henaff2019data,oord2018representation,tian2019contrastive,hjelm2018learning,chen2020simple,he2019momentum}. The common idea in these works is the following: pull together an anchor and a ``positive" sample in embedding space, and push apart the anchor from many ``negative" samples. Since no labels are available, a positive pair often consists of data augmentations of the sample, and negative pairs are formed by the anchor and randomly chosen samples from the minibatch. This is depicted in Fig. \ref{fig:teaser1} (left). In \cite{oord2018representation,tian2019contrastive}, connections are made of the contrastive loss to maximization of mutual information between different views of the data.

In this work, we propose a loss for supervised learning that builds on the contrastive self-supervised literature by leveraging label information.
Normalized embeddings from the \emph{same class} are pulled closer together than embeddings from \emph{different classes}. Our technical novelty in this work is to consider \emph{many positives} per anchor in addition to many negatives (as opposed to self-supervised contrastive learning which uses only a single positive). These positives are drawn from samples of the same class as the anchor, rather than being data augmentations of the anchor, as done in self-supervised learning. While this is a simple extension to the self-supervised setup, it is non-obvious how to setup the loss function correctly, and we analyze two alternatives. Fig.~\ref{fig:teaser1} (right) and Fig.~1 (Supplementary) provide a visual explanation of our proposed loss. Our loss can be seen as a generalization of both the triplet \cite{weinberger2009distance} and N-pair losses \cite{sohn2016improved}; the former uses only one positive and one negative sample per anchor, and the latter uses one positive and many negatives. The use of many positives and many negatives for each anchor allows us to achieve state of the art performance without the need for hard negative mining, which can be difficult to tune properly. To the best of our knowledge, this is the first contrastive loss to consistently perform better than cross-entropy on large-scale classification problems. Furthermore, it provides a unifying loss function that can be used for either self-supervised or supervised learning.

Our resulting loss, SupCon, is simple to implement and stable to train, as our empirical results show. It achieves excellent top-1 accuracy on the ImageNet dataset on the ResNet-50 and ResNet-200 architectures \cite{he2016deep}. On ResNet-200 \cite{cubuk2019autoaugment}, we achieve a top-1 accuracy of $81.4\%$, which is a $0.8\%$ improvement over the state of the art \cite{lim2019fast} cross-entropy loss on the same architecture (see Fig. \ref{fig:imagenet_top1_teaser}). The gain in top-1 accuracy is accompanied by increased robustness as measured on the ImageNet-C dataset \cite{hendrycks2019benchmarking}. Our main contributions are summarized below:

\begin{enumerate}[leftmargin=*,topsep=0pt,itemsep=-1ex,partopsep=1ex,parsep=1ex]
    \item We propose a novel extension to the contrastive loss function that allows for multiple positives per anchor, thus adapting contrastive learning to the fully supervised setting. Analytically and empirically, we show that a na{\"i}ve extension performs much worse than our proposed version.
    \item We show that our loss provides consistent boosts in top-1 accuracy for a number of datasets. It is also more robust to natural corruptions.
    \item We demonstrate analytically that the gradient of our loss function encourages learning from hard positives and hard negatives. %
    \item We show empirically that our loss is less sensitive than cross-entropy to a range of hyperparameters. %
\end{enumerate}
\vspace{-10pt}

\if{false}
Self-supervised learning is a branch of unsupervised learning that relies on natural supervision cues, usually exploiting structure in data. Examples include co-occurring modalities \cite{sun2019videobert,arandjelovic2017look,tian2019contrastive}, spatio-temporal structure \cite{henaff2019data,devlin2018bert} or generated structure \cite{hjelm2018learning}. The structure provides a form of weak supervision which can be exploited to learn powerful embeddings.  The resulting embeddings can be used for downstream tasks such as classification or object detection. Self-supervised learning has seen significant strides in recent years in multiple domains such as images, videos and natural language. 

The basic form of the loss function is inspired by noise contrastive estimation \cite{gutmann2010noise} and N-pair losses \cite{sohn2016improved}, which allows learning of unnormalized statistical models by training a binary classifier to distinguish between samples from noise and data distributions. In contrastive learning as used in the current paper and related work, we also employ a classification like approach. However, here the idea is to increase the similarity between positive pairs of samples and decrease similarity between negative pairs. Similarity is usually defined as the inner product between low-dimensional embeddings. The resulting embeddings are then shown to be a very good representation by being trained on various downstream transfer tasks. 

In addition to excellent empirical performance, theoretical motivation for unsupervised contrastive learning has been provided by \cite{arora2019theoretical}, where it is shown that the contrastive loss is an upper bound on downstream supervised loss, under the assumption that the data distribution is stationary between pre-training and downstream stages. Furthermore \cite{arora2019theoretical} suggest a novel loss that incorporates multiple positive samples, compared to the usual loss that contrasts a single positive with many negatives. The use of multiple positives provides a tighter bound on the downstream supervised loss. Our new loss function is closely related to this idea.

The family of \emph{metric losses} are intimately related to the contrastive loss. The canonical example is the triplet loss function  \cite{weinberger2009distance}, which introduced the idea of decreasing the embedding distance between positive pairs of samples, and push apart negative pairs. The triplet loss has proven very powerful in learning high performing embeddings e.g. \cite{schroff2015facenet,chopra2005learning}, and has often been used when the total number of categories are unknown, but it is known whether or not two samples are of the same category. The contrastive loss as used in this paper is very similar to the metric loss when we use a single positive and negative example for each anchor sample. However, the use of larger numbers of positives and negatives are very important for performance.

\fi
\section{Related Work}
\label{sec:related}
Our work draws on existing literature in self-supervised representation learning, metric learning and supervised learning. Here we focus on the most relevant papers. The cross-entropy loss was introduced as a powerful loss function to train deep networks \cite{rumelhart1986learning,baum1988supervised,levin1988accelerated}. The key idea is simple and intuitive: each class is assigned a target (usually 1-hot) vector. However, it is unclear why these target labels should be the optimal ones and some work has tried to identify better target label vectors, e.g. \cite{yang2015deep}. A number of papers have studied other drawbacks of the cross-entropy loss, such as sensitivity to noisy labels \cite{zhang2018generalized,sukhbaatar2014training}, presence of adversarial examples \cite{elsayed2018large,nar2019cross}, and poor margins \cite{cao2019learning}. Alternative losses have been proposed, but the most effective ideas in practice have been approaches that change the reference label distribution, such as label smoothing \cite{szegedy2016rethinking,muller2019does}, data augmentations such as Mixup \cite{zhang2017mixup} and CutMix \cite{yun2019cutmix}, and knowledge distillation \cite{hinton2015distilling}.

Powerful self-supervised representation learning approaches based on deep learning models have recently been developed in the natural language domain \cite{devlin2018bert,yang2019xlnet,mikolov2013distributed}. In the image domain, pixel-predictive approaches have also been used to learn embeddings \cite{doersch2015unsupervised,zhang2016colorful,zhang2017split,noroozi2016unsupervised}. These methods try to predict missing parts of the input signal. However, a more effective approach has been to replace a dense per-pixel predictive loss, with a loss in lower-dimensional representation space. The state of the art family of models for self-supervised representation learning using this paradigm are collected under the umbrella of contrastive learning \cite{wu2018unsupervised,henaff2019data,hjelm2018learning,tian2019contrastive,sermanet2017time,chen2020simple,tschannen2019mutual}. In these works, the losses are inspired by noise contrastive estimation \cite{gutmann2010noise,mnih2013learning} or N-pair losses \cite{sohn2016improved}. Typically, the loss is applied at the last layer of a deep network. At test time, the embeddings from a previous layer are utilized for downstream transfer tasks, fine tuning or direct retrieval tasks. \cite{he2019momentum} introduces the approximation of only back-propagating through part of the loss, and also the approximation of using stale representations in the form of a memory bank.

Closely related to contrastive learning is the family of losses based on metric distance learning or triplets \cite{chopra2005learning,weinberger2009distance,schroff2015facenet}. These losses have been used to learn powerful representations, often in supervised settings, where labels are used to guide the choice of positive and negative pairs. The key distinction between triplet losses and contrastive losses is the number of positive and negative pairs per data point; triplet losses use exactly one positive and one negative pair per anchor. In the supervised metric learning setting, the positive pair is chosen from the same class and the negative pair is chosen from other classes, nearly always requiring hard-negative mining for good performance \cite{schroff2015facenet}. Self-supervised contrastive losses similarly use just one positive pair for each anchor sample, selected using either co-occurrence \cite{henaff2019data,hjelm2018learning,tian2019contrastive} or data augmentation \cite{chen2020simple}. The major difference is that many negative pairs are used for each anchor. These are usually chosen uniformly at random using some form of weak knowledge, such as patches from other images, or frames from other randomly chosen videos, relying on the assumption that this approach yields a very low probability of false negatives.

 Resembling our supervised contrastive approach is the soft-nearest neighbors loss introduced in  \cite{salakhutdinov2007learning} and used in \cite{wu2018improving}. Like \cite{wu2018improving}, we improve upon \cite{salakhutdinov2007learning} by normalizing the embeddings and replacing euclidean distance with inner products. We further improve on \cite{wu2018improving} by the increased use of data augmentation, a disposable contrastive head and two-stage training (contrastive followed by cross-entropy), and crucially, changing the form of the loss function to significantly improve results (see Section 3). \cite{frosst2019analyzing} also uses a closely related loss formulation to ours to \textit{entangle} representations at intermediate layers by maximizing the loss. Most similar to our method is the Compact Clustering via Label Propagation (CCLP) regularizer in Kamnitsas et. al. \cite{Kamnitsas2018SemiSupervisedLV}. While CCLP focuses mostly on the semi-supervised case, in the fully supervised case the regularizer reduces to almost exactly our loss formulation. Important practical differences include our normalization of the contrastive embedding onto the unit sphere, tuning of a temperature parameter in the contrastive objective, and stronger augmentation. Additionally, Kamnitsas et. al. use the contrastive embedding as an input to a classification head, which is trained jointly with the CCLP regularizer, while SupCon employs a two stage training and discards the contrastive head. Lastly, the scale of experiments in Kamnitsas et. al. is much smaller than in this work. Merging the findings of our paper and CCLP is a promising direction for semi-supervised learning research.

\section{Method}

Our method is structurally similar to that used in \cite{tian2019contrastive,chen2020simple} for self-supervised contrastive learning, with modifications for supervised classification. Given an input batch of data, we first apply data augmentation twice to obtain two copies of the batch. Both copies are forward propagated through the encoder network to obtain a 2048-dimensional normalized embedding. During training, this representation is further propagated through a projection network that is discarded at inference time. The supervised contrastive loss is computed on the outputs of the projection network. To use the trained model for classification, we train a linear classifier on top of the frozen representations using a cross-entropy loss. Fig. 1 in the Supplementary material provides a visual explanation.

\subsection{Representation Learning Framework}
The main components of our framework are:
\begin{itemize}[leftmargin=*]
  \item [$\bullet$] \emph{Data Augmentation} module, $Aug(\cdot)$. For each input sample, $\boldsymbol{x}$, we generate two random augmentations, $\boldsymbol{\tilde{x}} = Aug(\boldsymbol{x})$,  each of which represents a different \emph{view} of the data and contains some subset of the information in the original sample. Sec. \ref{sec:experiments} gives details of the augmentations.

  \item[$\bullet$] \emph{Encoder Network}, $Enc(\cdot)$, which maps $\boldsymbol{x}$ to a representation vector, $\boldsymbol{r}=Enc(\boldsymbol{x})\in\mathcal{R}^{D_E}$. Both augmented samples are separately input to the same encoder, resulting in a pair of representation vectors. $\boldsymbol{r}$ is normalized to the unit hypersphere in $\mathcal{R}^{D_E}$ ($D_E = 2048$ in all our experiments in the paper). Consistent with the findings of \cite{schroff2015facenet,wang2020understanding}, our analysis and experiments show that this normalization improves top-1 accuracy. %
  
  \item[$\bullet$] \emph{Projection Network}, $Proj(\cdot)$, which maps $\boldsymbol{r}$ to a vector $\boldsymbol{z}=Proj(\boldsymbol{r})\in\mathcal{R}^{D_P}$. We instantiate $Proj(\cdot)$ as either a multi-layer perceptron \cite{hastie2001statisticallearning} with a single hidden layer of size $2048$ and output vector of size $D_P=128$ or just a single linear layer of size $D_P=128$; we leave to future work the investigation of optimal $Proj(\cdot)$ architectures. We again normalize the output of this network to lie on the unit hypersphere, which enables using an inner product to measure distances in the projection space. As in self-supervised contrastive learning \cite{tian2019contrastive,chen2020simple}, we discard $Proj(\cdot)$ at the end of contrastive training. As a result, our inference-time models contain exactly the same number of parameters as a cross-entropy model using the same encoder, $Enc(\cdot)$. 
\end{itemize}

\subsection{Contrastive Loss Functions}
\label{sec:contrastive_losses}
Given this framework, we now look at the family of contrastive losses, starting from the self-supervised domain and analyzing the options for adapting it to the supervised domain, showing that one formulation is superior. For a set of $N$ randomly sampled sample/label pairs, $\{\boldsymbol{x}_k,\boldsymbol{y}_k\}_{k=1...N}$, the corresponding batch used for training consists of $2N$ pairs, $\{\boldsymbol{\tilde{x}}_\ell,\boldsymbol{\tilde{y}}_\ell\}_{\ell=1...2N}$, where $\boldsymbol{\tilde{x}}_{2k}$ and $\boldsymbol{\tilde{x}}_{2k-1}$ are two random augmentations (a.k.a., ``views") of $\boldsymbol{x}_k$ ($k=1...N$) and $\boldsymbol{\tilde{y}}_{2k-1}=\boldsymbol{\tilde{y}}_{2k}=\boldsymbol{y}_k$. For the remainder of this paper, we will refer to a set of $N$ samples as a ``batch" and the set of $2N$ augmented samples as a ``multiviewed batch".

\subsubsection{Self-Supervised Contrastive Loss}
Within a multiviewed batch, let $i\in I\equiv\{1...2N\}$ be the index of an arbitrary augmented sample, and let $j(i)$ be the index of the other augmented sample originating from the same source sample. In \emph{self-supervised} contrastive learning (e.g., \cite{chen2020simple,tian2019contrastive,henaff2019data,hjelm2018learning}), the loss takes the following form.
\begin{equation}
  \mathcal{L}^{self}
  =\sum_{i\in I}\mathcal{L}_i^{self}
  =-\sum_{i\in I}\log{
  \frac{\text{exp}\left(\boldsymbol{z}_i\bigcdot\boldsymbol{z}_{j(i)}/\tau\right)}{\sum\limits_{a\in A(i)}\text{exp}\left(\boldsymbol{z}_i\bigcdot\boldsymbol{z}_a/\tau\right)}
  }
  \label{eqn:self_loss}
\end{equation}
Here, $\boldsymbol{z}_\ell=Proj(Enc(\boldsymbol{\tilde{x}}_\ell))\in\mathcal{R}^{D_P}$, the $\bigcdot$ symbol denotes the inner (dot) product, $\tau\in\mathcal{R}^+$ is a scalar temperature parameter, and $A(i)\equiv I\setminus\{i\}$. The index $i$ is called the \emph{anchor}, index $j(i)$ is called the \emph{positive}, and the other $2(N-1)$ indices ($\{k\in A(i)\setminus\{j(i)\}$) are called the \emph{negatives}. Note that for each anchor $i$, there is $1$ positive pair and $2N - 2$ negative pairs. The denominator has a total of $2N - 1$ terms (the positive and negatives). 

\subsubsection{Supervised Contrastive Losses}
For supervised learning, the contrastive loss in Eq. \ref{eqn:self_loss} is incapable of handling the case where, due to the presence of labels, more than one sample is known to belong to the same class. Generalization to an arbitrary numbers of positives, though, leads to a choice between multiple possible functions. Eqs. \ref{eqn:supervised_loss} and \ref{eqn:bad_supervised_loss} present the two most straightforward ways to generalize Eq. \ref{eqn:self_loss} to incorporate supervision.
\begin{equation}
  \mathcal{L}_{out}^{sup}
  =\sum_{i\in I}\mathcal{L}_{out,i}^{sup}
  =\sum_{i\in I}\frac{-1}{|P(i)|}\sum_{p\in P(i)}\log{\frac{\text{exp}\left(\boldsymbol{z}_i\bigcdot\boldsymbol{z}_p/\tau\right)}{\sum\limits_{a\in A(i)}\text{exp}\left(\boldsymbol{z}_i\bigcdot\boldsymbol{z}_a/\tau\right)}}
  \label{eqn:supervised_loss}
\end{equation}
\begin{equation}
  \mathcal{L}_{in}^{sup}
  =\sum_{i\in I}\mathcal{L}_{in,i}^{sup}
  =\sum_{i\in I}-\log\left\{\frac{1}{|P(i)|}\sum_{p\in P(i)}\frac{\text{exp}\left(\boldsymbol{z}_i\bigcdot\boldsymbol{z}_p/\tau\right)}{\sum\limits_{a\in A(i)}\text{exp}\left(\boldsymbol{z}_i\bigcdot\boldsymbol{z}_a/\tau\right)}\right\}
  \label{eqn:bad_supervised_loss}
\end{equation}
Here, $P(i)\equiv\{p\in A(i):\boldsymbol{\tilde{y}}_p=\boldsymbol{\tilde{y}}_i\}$ is the set of indices of all positives in the multiviewed batch distinct from $i$, and $|P(i)|$ is its cardinality. In Eq. \ref{eqn:supervised_loss}, the summation over positives is located \emph{outside} of the $\log$ ($\mathcal{L}_{out}^{sup}$) while in Eq. \ref{eqn:bad_supervised_loss}, the summation is located \emph{inside} of the $\log$ ($\mathcal{L}_{in}^{sup}$). Both losses have the following desirable properties:

\begin{itemize}[leftmargin=*]
  \item[$\bullet$] \textbf{Generalization to an arbitrary number of positives.} The major structural change of Eqs. \ref{eqn:supervised_loss} and \ref{eqn:bad_supervised_loss} over Eq. \ref{eqn:self_loss} is that now, for any anchor, \emph{all} positives in a multiviewed batch (i.e., the augmentation-based sample as well as any of the remaining samples with the same label) contribute to the numerator. For randomly-generated batches whose size is large with respect to the number of classes, multiple additional terms will be present (on average, $N/C$, where $C$ is the number of classes). The supervised losses encourage the encoder to give closely aligned representations to \emph{all} entries from the same class, resulting in a more robust clustering of the representation space than that generated from Eq. \ref{eqn:self_loss}, as is supported by our experiments in Sec.~\ref{sec:experiments}. \if{false}{In Fig.~\ref{fig:teaser}(left), we show this visually for the case of $2$ classes.}\fi

  \item[$\bullet$] \textbf{Contrastive power increases with more negatives.} Eqs. \ref{eqn:supervised_loss} and \ref{eqn:bad_supervised_loss} both preserve the summation over negatives in the contrastive denominator of Eq. \ref{eqn:self_loss}. This form is largely motivated by noise contrastive estimation and N-pair losses \cite{gutmann2010noise,sohn2016improved}, wherein the ability to discriminate between signal and noise (negatives) is improved by adding more examples of negatives. This property is important for representation learning via self-supervised contrastive learning, with many papers showing increased performance with increasing number of negatives \cite{henaff2019data,he2019momentum,tian2019contrastive,chen2020simple}.

  \item[$\bullet$] \textbf{Intrinsic ability to perform hard positive/negative mining.} When used with \emph{normalized} representations, the loss in Eq. \ref{eqn:self_loss} induces a gradient structure that gives rise to implicit hard positive/negative mining. The gradient contributions from \emph{hard} positives/negatives (i.e., ones against which continuing to contrast the anchor \emph{greatly} benefits the encoder) are large while those for \emph{easy} positives/negatives (i.e., ones against which continuing to contrast the anchor only \emph{weakly} benefits the encoder) are small. Furthermore, for hard positives, the effect increases (asymptotically) as the number of negatives does. Eqs. \ref{eqn:supervised_loss} and \ref{eqn:bad_supervised_loss} both preserve this useful property and generalize it to all positives. This implicit property allows the contrastive loss to sidestep the need for explicit hard mining, which is a delicate but critical part of many losses, such as triplet loss \cite{schroff2015facenet}. We note that this implicit property applies to both supervised and self-supervised contrastive losses, but our derivation is the first to clearly show this property. We provide a full derivation of this property from the loss gradient in the Supplementary material.
\end{itemize}

\begin{wraptable}{r}{0.35\textwidth}
    \vspace{-20pt}
    \centering
  \centering
  \begin{tabular}{cc}
    \toprule
    Loss & Top-1 \\
    \midrule
    &\\[-.9em]
    $\mathcal{L}_{out}^{sup}$ & 78.7\% \\
    $\mathcal{L}_{in}^{sup}$ & 67.4\% \\
    \bottomrule
  \end{tabular}
  \caption{\small ImageNet Top-1 classification accuracy for supervised contrastive losses on ResNet-50 for a batch size of 6144. %
  }
  \vspace{-10pt}
  \label{tab:supervised_loss_variants}
\end{wraptable}

The two loss formulations are not, however, equivalent.
Because $\log$ is a concave function, Jensen's Inequality \cite{jensen1906sur} implies that $\mathcal{L}_{in}^{sup}\le\mathcal{L}_{out}^{sup}$. One would thus expect $\mathcal{L}_{out}^{sup}$ to be the superior supervised loss function (since it upper-bounds $\mathcal{L}_{in}^{sup}$). This conclusion is also supported analytically.
Table \ref{tab:supervised_loss_variants} compares the ImageNet \cite{deng2009imagenet} top-1 classification accuracy using $\mathcal{L}_{out}^{sup}$ and $\mathcal{L}_{in}^{sup}$ for different batch sizes ($N$) on the ResNet-50 \cite{he2016deep} architecture. The $\mathcal{L}_{out}^{sup}$ supervised loss achieves significantly higher performance than $\mathcal{L}_{in}^{sup}$. We conjecture that this is due to the gradient of $\mathcal{L}_{in}^{sup}$ having structure less optimal for training than that of $\mathcal{L}_{out}^{sup}$. For $\mathcal{L}_{out}^{sup}$, the positives normalization factor (i.e., $1/|P(i)|$) serves to remove bias present in the positives in a multiviewed batch contributing to the loss. However, though $\mathcal{L}_{in}^{sup}$ also contains the same normalization factor, it is located \emph{inside} of the $\log$. It thus contributes only an additive constant to the overall loss, which does not affect the gradient. Without any normalization effects, the gradients of $\mathcal{L}_{in}^{sup}$ are more susceptible to bias in the positives, leading to sub-optimal training.

An analysis of the gradients themselves supports this conclusion. As shown in the Supplementary, the gradient for \emph{either} $\mathcal{L}_{out,i}^{sup}$ or $\mathcal{L}_{in,i}^{sup}$ with respect to the embedding $\boldsymbol{z}_i$ has the following form. %
\begin{equation}
  \label{eqn:gradient}
  \frac{\partial\mathcal{L}_i^{sup}}{\partial\boldsymbol{z}_i} = \frac{1}{\tau}\left\{
  \sum_{p\in P(i)}\boldsymbol{z}_p(P_{ip}-X_{ip})+
  \sum_{n\in N(i)}\boldsymbol{z}_nP_{in}
  \right\}
\end{equation}
Here, $N(i)\equiv\{n\in A(i):\boldsymbol{\tilde{y}}_n\neq\boldsymbol{\tilde{y}}_i\}$ is the set of indices of all negatives in the multiviewed batch, and $P_{ix}\equiv\text{exp}\left(\boldsymbol{z}_i\bigcdot\boldsymbol{z}_x/\tau\right)/\sum_{a\in A(i)}\text{exp}\left(\boldsymbol{z}_i\bigcdot\boldsymbol{z}_a/\tau\right)$. The difference between the gradients for the two losses is in $X_{ip}$.
\begin{equation}
  X_{ip}=\left\{
  \begin{matrix}
  \frac{\text{exp}\left(\boldsymbol{z}_i\bigcdot\boldsymbol{z}_p/\tau\right)}{\sum\limits_{p'\in P(i)}\text{exp}\left(\boldsymbol{z}_i\bigcdot\boldsymbol{z}_{p'}/\tau\right)} & , & \text{if }\mathcal{L}_i^{sup}=\mathcal{L}_{in,i}^{sup}\\
  \frac{1}{|P(i)|} & , & \text{if }\mathcal{L}_i^{sup}=\mathcal{L}_{out,i}^{sup}
  \end{matrix}\right.
\end{equation}
If each $\boldsymbol{z}_p$ is set to the (less biased) mean positive representation vector, $\overline{\boldsymbol{z}}$, $X_{ip}^{in}$ reduces to $X_{ip}^{out}$:
\begin{equation}
  \left.X_{ip}^{in}\right|_{\boldsymbol{z}_p=\overline{\boldsymbol{z}}}
  =\frac{\text{exp}\left(\boldsymbol{z}_i\bigcdot\overline{\boldsymbol{z}}/\tau\right)}{\sum\limits_{p'\in P(i)}\text{exp}\left(\boldsymbol{z}_i\bigcdot\overline{\boldsymbol{z}}/\tau\right)}
  =\frac{\text{exp}\left(\boldsymbol{z}_i\bigcdot\overline{\boldsymbol{z}}/\tau\right)}{|P(i)|\cdot\text{exp}\left(\boldsymbol{z}_i\bigcdot\overline{\boldsymbol{z}}/\tau\right)}
  =\frac{1}{|P(i)|}
  =X_{ip}^{out}
\end{equation}
From the form of $\partial\mathcal{L}_i^{sup}/\partial\boldsymbol{z}_i$, we conclude that the stabilization due to using the mean of positives benefits training. %
Throughout the rest of the paper, we consider only $\mathcal{L}_{out}^{sup}$.

\subsubsection{Connection to Triplet Loss and N-pairs Loss}
Supervised contrastive learning is closely related to the triplet loss \cite{weinberger2009distance}, one of the widely-used loss functions for supervised learning. In the Supplementary, we show that the triplet loss is a special case of the contrastive loss when one positive and one negative are used. When more than one negative is used, we show that the SupCon loss becomes equivalent to the N-pairs loss \cite{sohn2016improved}.

\section{Experiments}
\label{sec:experiments}
We evaluate our SupCon loss ($\mathcal{L}_{out}^{sup}$, Eq. \ref{eqn:supervised_loss}) by measuring classification accuracy on a number of common image classification benchmarks including CIFAR-10 and CIFAR-100 \cite{krizhevsky2009learning} and ImageNet \cite{deng2009imagenet}. We also benchmark our ImageNet models on robustness to common image corruptions \cite{hendrycks2019benchmarking} and show how performance varies with changes to hyperparameters and reduced data. For the encoder network ($Enc(\cdot)$) we experimented with three commonly used encoder architectures: ResNet-50, ResNet-101, and ResNet-200 \cite{he2016deep}. The normalized activations of the final pooling layer ($D_E=2048$) are used as the representation vector. We experimented with four different implementations of the $Aug(\cdot)$ data augmentation module: AutoAugment \cite{cubuk2019autoaugment}; RandAugment \cite{cubuk2019randaugment}; SimAugment \cite{chen2020simple}, and Stacked RandAugment \cite{tian2020makes} (see details of our SimAugment and Stacked RandAugment implementations in the Supplementary). AutoAugment outperforms all other data augmentation strategies on ResNet-50 for both SupCon and cross-entropy. Stacked RandAugment performed best for ResNet-200 for both loss functions. We provide more details in the Supplementary.

\subsection{Classification Accuracy}
Table \ref{tab:datasets} shows that SupCon generalizes better than cross-entropy, margin classifiers (with use of labels) and unsupervised contrastive learning techniques on CIFAR-10, CIFAR-100 and ImageNet datasets. Table \ref{table:imagenet_top1} shows results for ResNet-50 and ResNet-200 (we use ResNet-v1 \cite{he2016deep}) for ImageNet. We achieve a new state of the art accuracy of $78.7\%$ on ResNet-50 with AutoAugment (for comparison, a number of the other top-performing methods are shown in Fig. \ref{fig:imagenet_top1_teaser}). Note that we also achieve a slight improvement over CutMix \cite{yun2019cutmix}, which is considered to be a state of the art data augmentation strategy. Incorporating data augmentation strategies such as CutMix \cite{yun2019cutmix} and MixUp \cite{zhang2017mixup} into contrastive learning could potentially improve results further. 

We also experimented with memory based alternatives \cite{he2019momentum}. On ImageNet, with a memory size of 8192 (requiring only the storage of 128-dimensional vectors), a batch size of 256, and SGD optimizer, running on 8 Nvidia V100 GPUs, SupCon is able to achieve $79.1\%$ top-1 accuracy on ResNet-50. This is in fact slightly better than the $78.7\%$ accuracy with 6144 batch size (and no memory);
and with significantly reduced compute and memory footprint.

Since SupCon uses 2 views per sample, its batch sizes are effectively twice the cross-entropy equivalent. We therefore also experimented with the cross-entropy ResNet-50 baselines using a batch size of 12,288. These only achieved $77.5\%$ top-1 accuracy. We additionally experimented with increasing the number of training epochs for cross-entropy all the way to 1400, but this actually decreased accuracy ($77.0\%$).

We tested the N-pairs loss \cite{sohn2016improved} in our framework with a batch size of 6144. N-pairs achieves only $57.4\%$ top-1 accuracy on ImageNet. We believe this is due to multiple factors missing from N-pairs loss compared to supervised contrastive: the use of multiple views; lower temperature; and many more positives. We show some results of the impact of the number of positives per anchor in the Supplementary (Sec. 6), and the N-pairs result is inline with them. We also note that the original N-pairs paper \cite{sohn2016improved} has already shown the outperformance of N-pairs loss to triplet loss. 

\begin{table}[t]
    \centering
     \vspace{-5pt}
    \begin{tabular}{ccccc}
        \toprule
        Dataset & SimCLR\cite{chen2020simple} & Cross-Entropy & Max-Margin \cite{liu2016largemargin}  &  SupCon \\\midrule
        CIFAR10  & 93.6 & 95.0 & 92.4 & \bf{96.0} \\
        CIFAR100 & 70.7 & 75.3 & 70.5 & \bf{76.5} \\
        ImageNet & 70.2 & 78.2 & 78.0 & \bf{78.7} \\
        \bottomrule
    \end{tabular}
    \vspace{2mm}
    \caption{Top-1 classification accuracy on ResNet-50 \cite{he2016deep} for various datasets. We compare cross-entropy training, unsupervised representation learning (SimCLR \cite{chen2020simple}), max-margin classifiers \cite{liu2016largemargin} and SupCon (ours). We re-implemented and tuned hyperparameters for all baseline numbers except margin classifiers where we report published results.  Note that the CIFAR-10 and CIFAR-100 results are from our PyTorch implementation and ImageNet from our TensorFlow implementation.}
    \vspace{-5pt}
    \label{tab:datasets}
\end{table}

\begin{table}[t]
 \vspace{-10pt}
\centering
\begin{tabular}{ccccc} 
 \toprule
  Loss & Architecture & Augmentation & Top-1 & Top-5 \\\midrule
 Cross-Entropy (baseline) & ResNet-50  & MixUp \cite{zhang2017mixup} & 77.4 & 93.6 \\
 Cross-Entropy (baseline) & ResNet-50 & CutMix \cite{yun2019cutmix} & 78.6 & 94.1 \\
 Cross-Entropy  (baseline) & ResNet-50 & AutoAugment \cite{cubuk2019autoaugment} & 78.2 & 92.9 \\
 Cross-Entropy (our impl.) & ResNet-50 & AutoAugment \cite{lim2019fast} & 77.6 & 95.3 \\ 
 SupCon & ResNet-50 & AutoAugment \cite{cubuk2019autoaugment}
 & {\bf 78.7} & {\bf 94.3} \\ \midrule
 Cross-Entropy (baseline) & ResNet-200 & AutoAugment \cite{cubuk2019autoaugment} & 80.6 & 95.3 \\ 
 Cross-Entropy (our impl.) & ResNet-200 & Stacked RandAugment \cite{tian2020makes} & 80.9 & 95.2 \\ 
 SupCon & ResNet-200 & Stacked RandAugment \cite{tian2020makes}
 & {\bf 81.4} & {\bf 95.9} \\ \midrule
 SupCon & ResNet-101 & Stacked RandAugment \cite{tian2020makes} & 80.2 & 94.7 \\ \bottomrule
 
\end{tabular}
\vspace{2mm}
\caption{Top-1/Top-5 accuracy results on ImageNet for AutoAugment \cite{cubuk2019autoaugment} with ResNet-50 and for Stacked RandAugment \cite{tian2020makes} with ResNet-101 and ResNet-200. The baseline numbers are taken from the referenced papers, and we also re-implement cross-entropy.}
 \vspace{-20pt}
\label{table:imagenet_top1}
\end{table}

\begin{figure}[t]
\centering
\begin{minipage}[t]{.6\linewidth}
\small
\setlength{\tabcolsep}{2pt}
\vspace{0pt}
\centering
\begin{tabular}{cccc} 
 \toprule
  Loss & Architecture & rel. mCE & mCE \\\midrule
 Cross-Entropy & AlexNet \cite{krizhevsky2012imagenet} & 100.0 & 100.0 \\
 (baselines)& VGG-19+BN \cite{Simonyan15} & 122.9 & 81.6 \\
 & ResNet-18 \cite{he2016deep} & 103.9 & 84.7 \\\midrule
 Cross-Entropy & ResNet-50 & 96.2 & 68.6 \\
 (our implementation)& ResNet-200 & 69.1 & 52.4 \\ 
 \midrule
 Supervised Contrastive & ResNet-50 & {\bf 94.6} & {\bf 67.2} \\
 & ResNet-200 & {\bf 66.5} & {\bf 50.6} \\ \bottomrule
\end{tabular}
\end{minipage}%
\begin{minipage}[t]{.4\linewidth}
\vspace{0pt}
\centering
\includegraphics[width=0.9\linewidth]{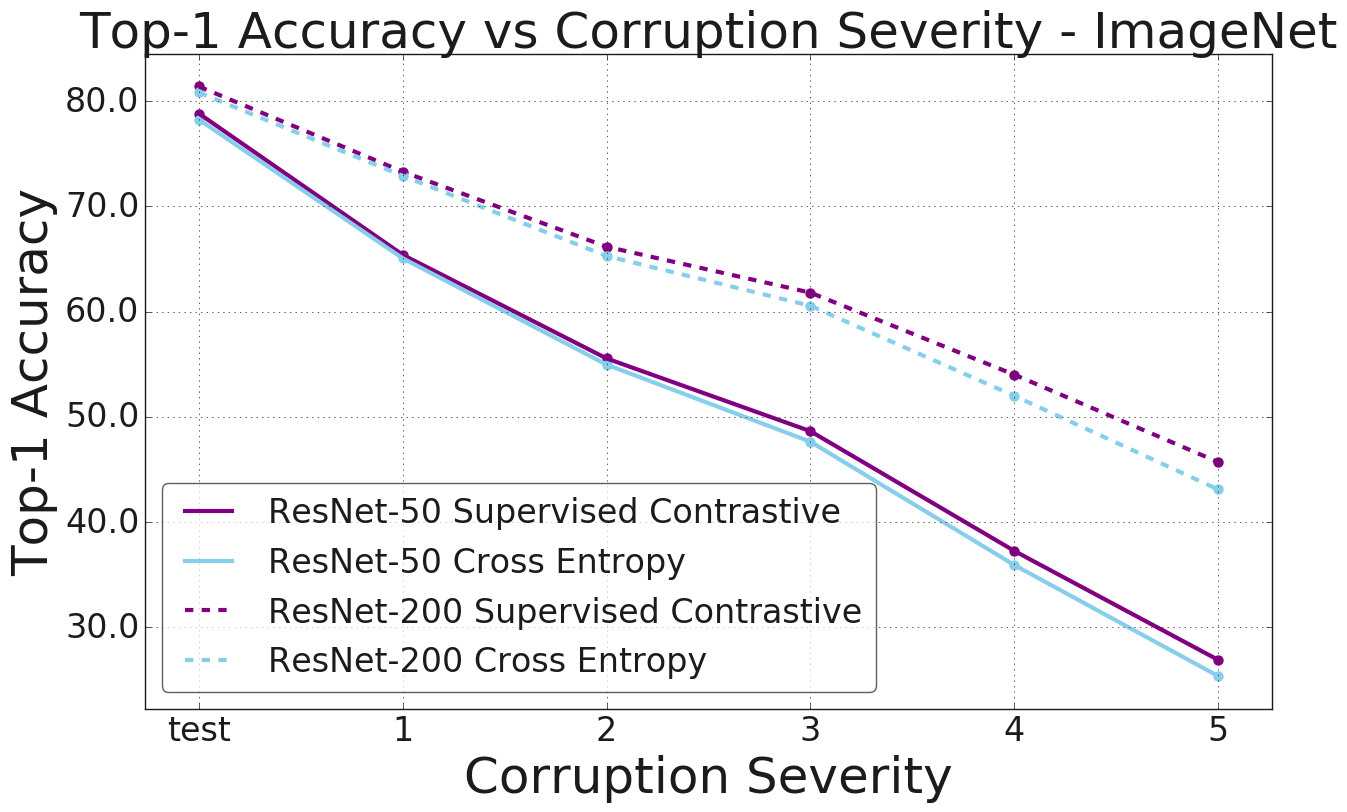}
\end{minipage}

\caption{ Training with supervised contrastive loss makes models more robust to corruptions in images. {\bf Left}: Robustness as measured by Mean Corruption Error (mCE) and relative mCE over the ImageNet-C dataset \cite{hendrycks2019benchmarking} (lower is better). {\bf Right}: Mean Accuracy as a function of corruption severity averaged over all various corruptions. (higher is better).}
\label{table:robustness}

\end{figure}

\subsection{Robustness to Image Corruptions and Reduced Training Data}
Deep neural networks lack robustness to out of distribution data or natural corruptions such as noise, blur and JPEG compression. The benchmark ImageNet-C dataset \cite{hendrycks2019benchmarking} is used to measure trained model performance on such corruptions. In Fig. \ref{table:robustness}(left), we compare the supervised contrastive models to cross-entropy using the Mean Corruption Error (mCE) and Relative Mean Corruption Error metrics \cite{hendrycks2019benchmarking}. Both metrics measure average degradation in performance compared to ImageNet test set, averaged over all possible corruptions and severity levels. Relative mCE is a better metric when we compare models with different Top-1 accuracy, while mCE is a better measure of absolute robustness to corruptions. The SupCon models have lower mCE values across different corruptions, showing increased robustness. We also see from Fig. \ref{table:robustness}(right) that SupCon models demonstrate lesser degradation in accuracy with increasing corruption severity.

\begin{figure}
    \centering
    \includegraphics[width=0.24\linewidth]{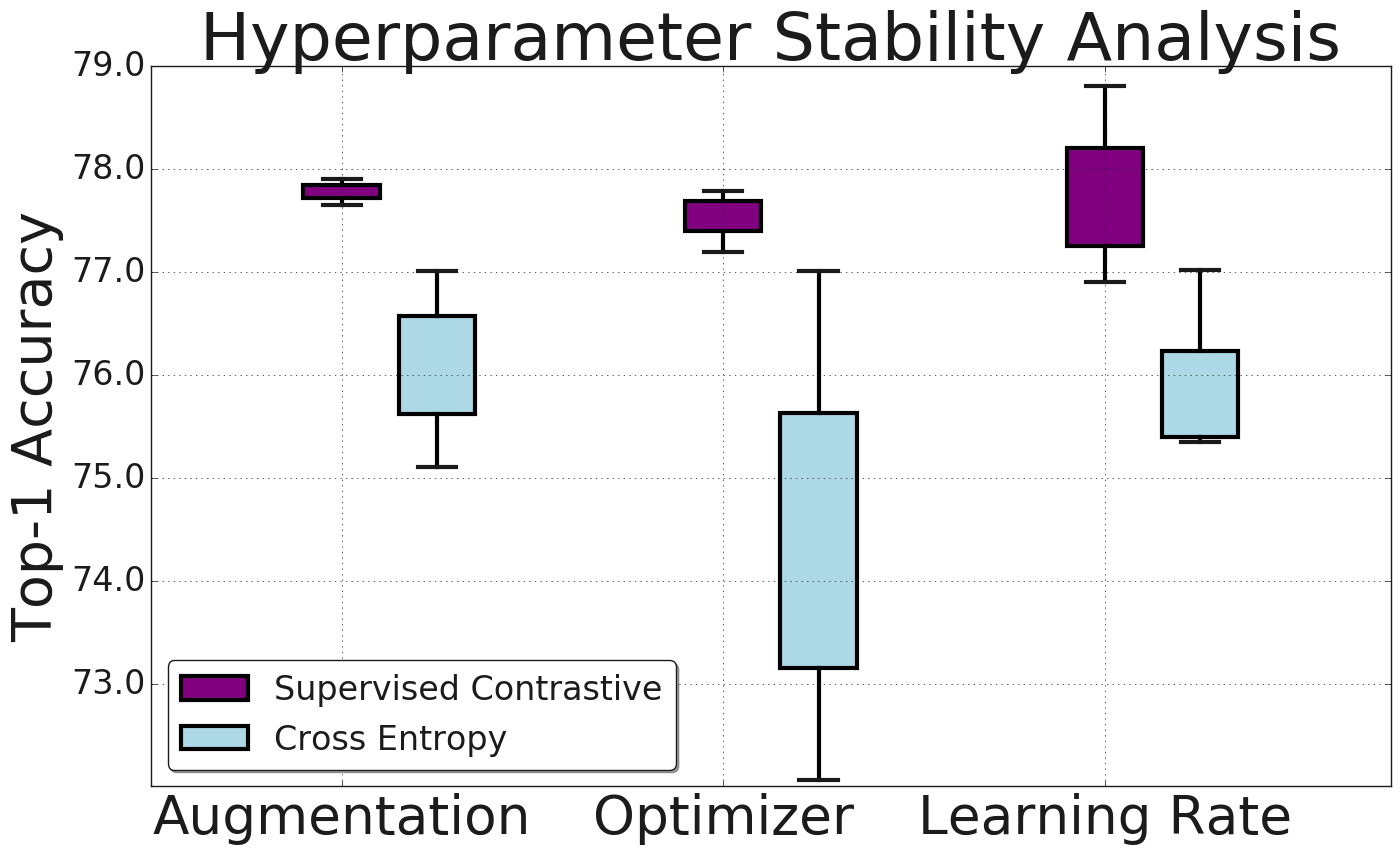}
    \includegraphics[width=0.24\linewidth]{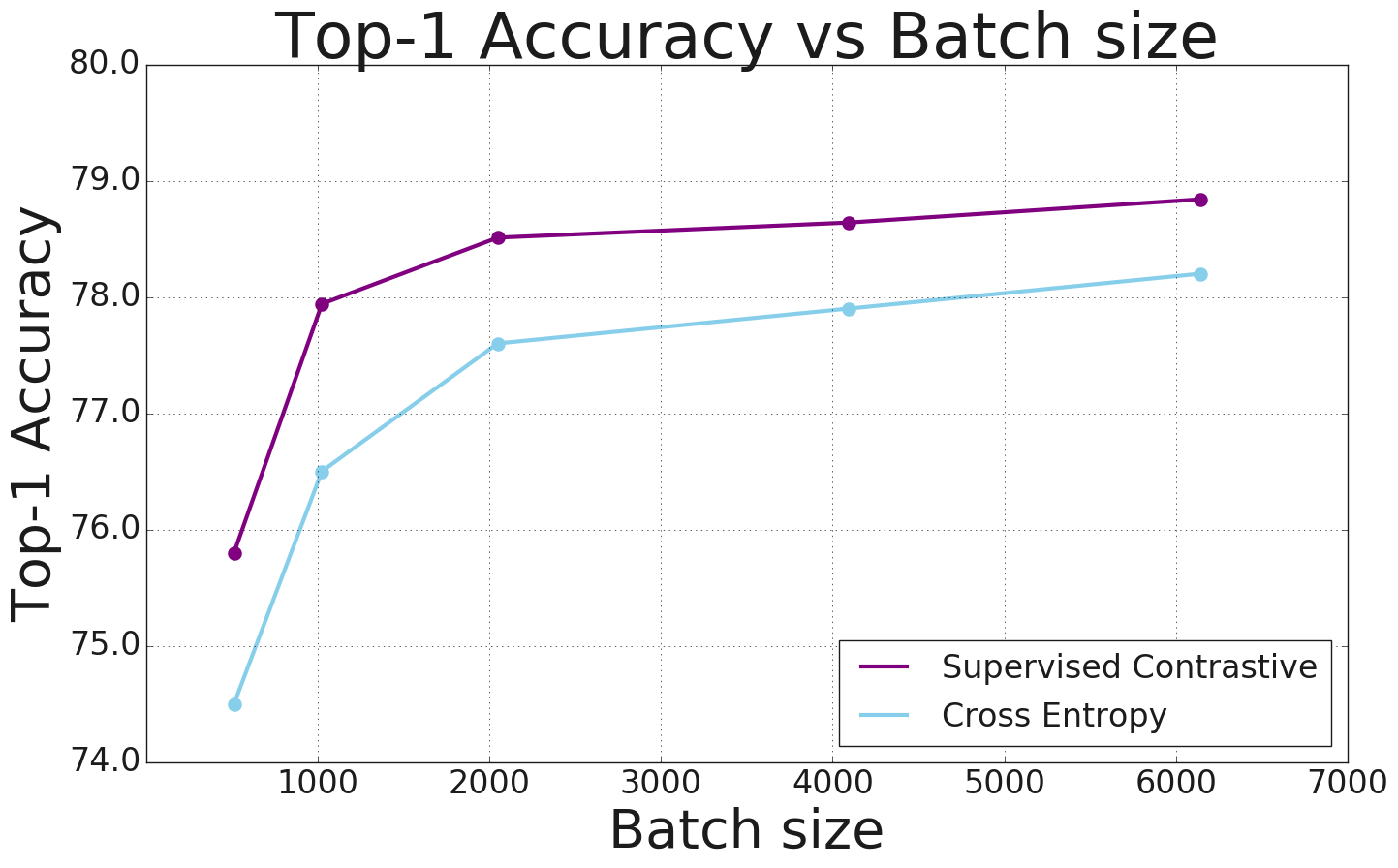}
    \includegraphics[width=0.24\linewidth]{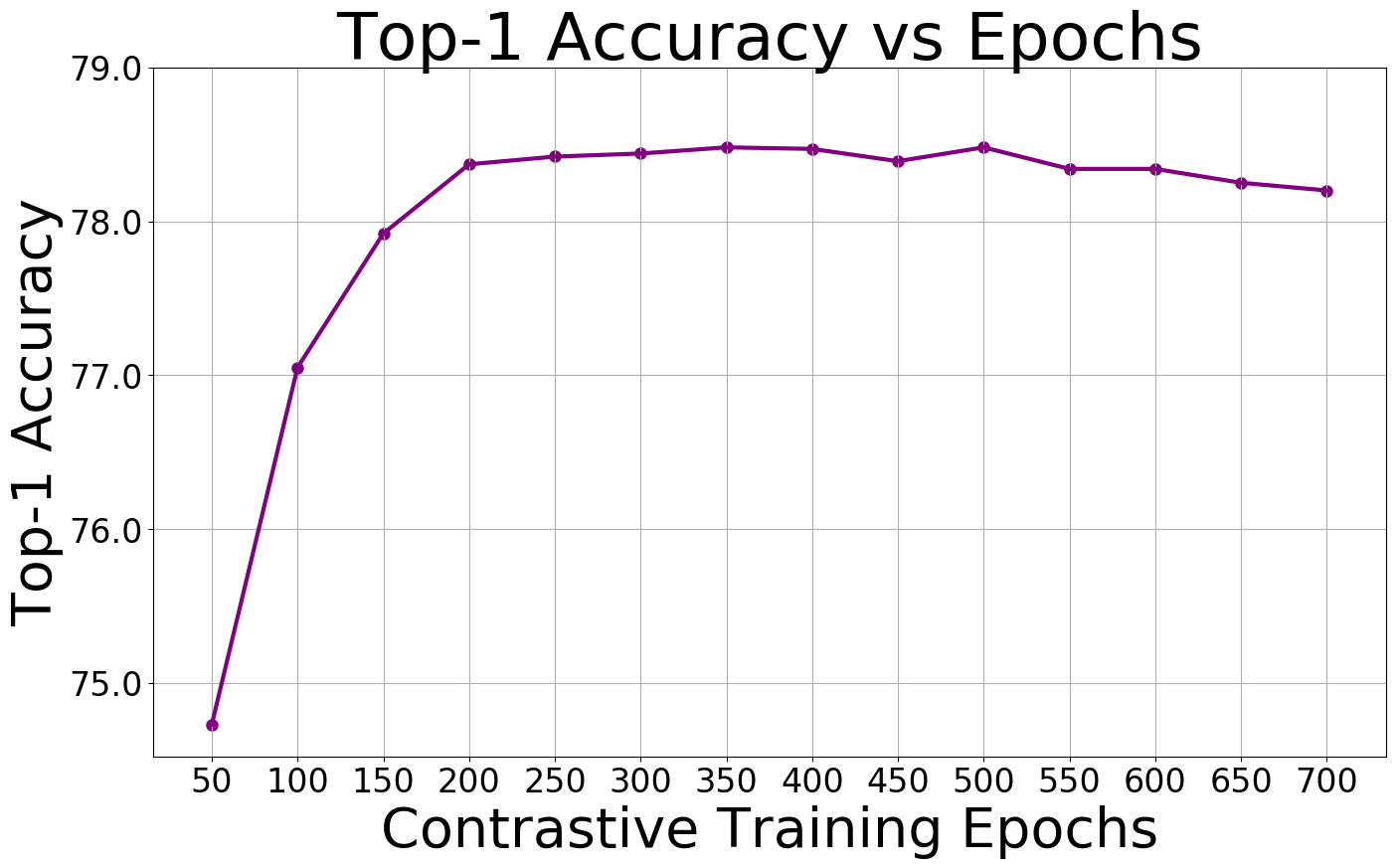}
    \includegraphics[width=0.24\linewidth]{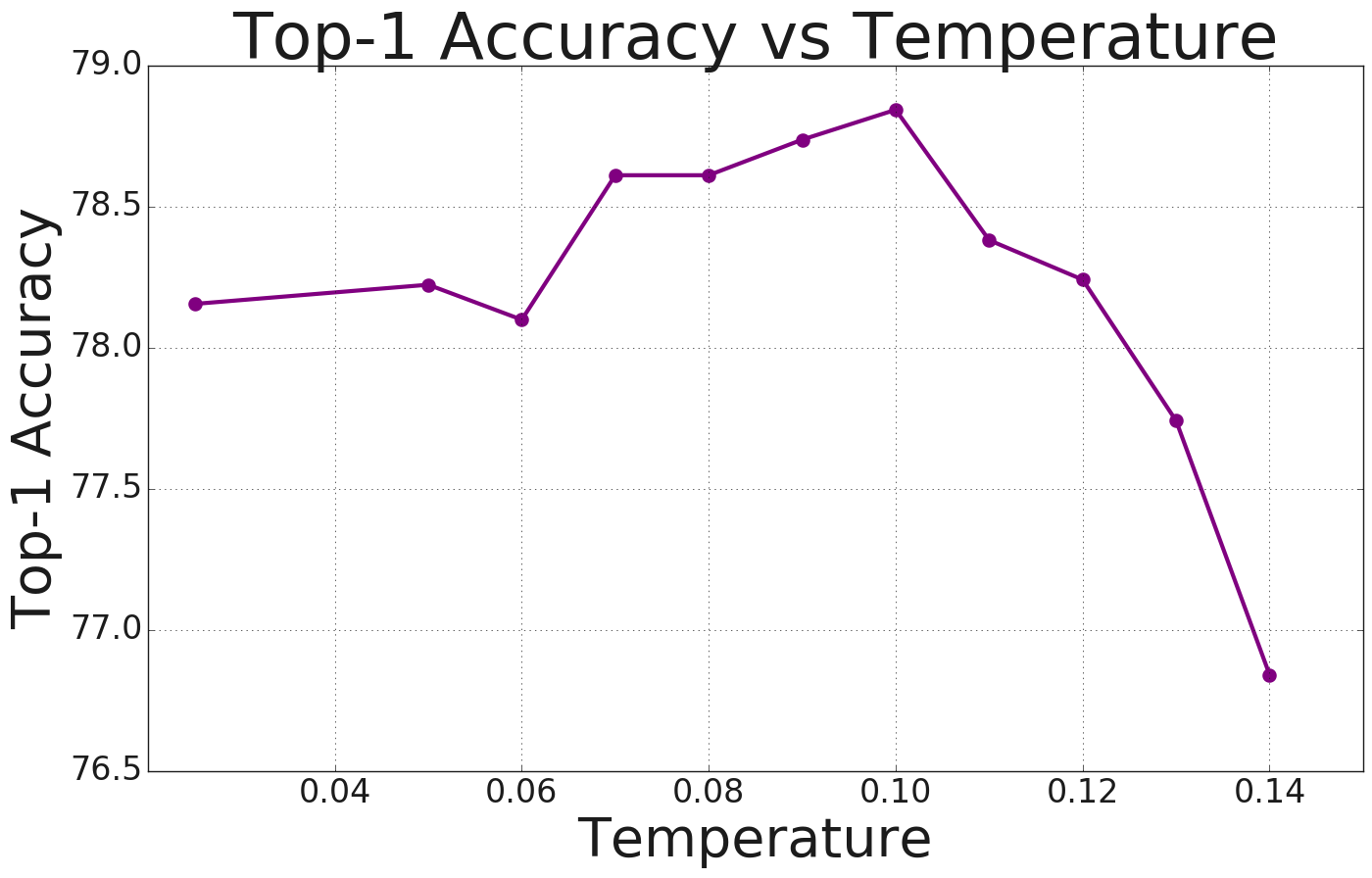}
    \caption{Accuracy of cross-entropy and supervised contrastive loss as a function of hyperparameters and training data size, all measured on ImageNet with a ResNet-50 encoder. (From left to right)
    {\bf (a)}: Standard boxplot showing Top-1 accuracy vs changes in augmentation, optimizer and learning rates. 
    {\bf (b)}: Top-1 accuracy as a function of batch size shows both losses benefit from larger batch sizes while Supervised Contrastive has higher Top-1 accuracy even when trained with smaller batch sizes. 
    {\bf (c)}: Top-1 accuracy as a function of SupCon pretraining epochs.
     {\bf (d)}: Top-1 accuracy as a function of temperature during pretraining stage for SupCon. 
     }
    \label{fig:hparam_stability} 
\end{figure}

\subsection{Hyperparameter Stability}
We experimented with hyperparameter stability by changing augmentations, optimizers and learning rates one at a time from the best combination for each of the methodologies. In Fig. \ref{fig:hparam_stability}(a), we compare the top-1 accuracy of SupCon loss against cross-entropy across changes in augmentations (RandAugment \cite{cubuk2019randaugment}, AutoAugment \cite{cubuk2019autoaugment}, SimAugment \cite{chen2020simple}, Stacked RandAugment \cite{tian2020makes}); optimizers (LARS, SGD with Momentum and RMSProp); and learning rates. We observe significantly lower variance in the output of the contrastive loss. Note that batch sizes for cross-entropy and supervised contrastive are the same, thus ruling out any batch-size effects. In Fig. \ref{fig:hparam_stability}(b), sweeping batch size and holding all other hyperparameters constant results in consistently better top-1 accuracy of the supervised contrastive loss.

\subsection{Transfer Learning}
We evaluate the learned representation for fine-tuning on 12 natural image datasets, following the protocol in Chen et.al. \cite{chen2020simple}. SupCon is on par with cross-entropy and \emph{self-supervised} contrastive loss on transfer learning performance when trained on the same architecture (Table \ref{table:transfer}). Our results are consistent with the findings in \cite{he2019rethinking} and \cite{kornblith2019better}: while better ImageNet models are correlated with better transfer performance, the dominant factor is architecture. Understanding the connection between training objective and transfer performance is left to future work. 

\begin{table}[t!]
\scriptsize
\setlength{\tabcolsep}{2pt}
\begin{tabular}{cccccccccccccc} 
 \toprule
   & Food & CIFAR10 & CIFAR100  & Birdsnap & SUN397 & Cars & Aircraft & VOC2007 & DTD & Pets & Caltech-101 & Flowers & Mean \\\midrule
 SimCLR-50 \cite{chen2020simple} &  \textbf{88.20} & \textbf{97.70} & \textbf{85.90} & \textbf{75.90} & \textbf{63.50} & 91.30 & \textbf{88.10} & 84.10 & 73.20 & 89.20 & 92.10 & \textbf{97.00} & \textbf{84.81}\\ 
 Xent-50 & 87.38 &	96.50 &	84.93 &	74.70 &	63.15 &	89.57 &	80.80 &	\textbf{85.36} &	\textbf{76.86} &	92.35 &	\textbf{92.34} &	96.93 & \textbf{84.67}\\
  SupCon-50 & 87.23 &	97.42 &	84.27 &	75.15 &	58.04 &	\textbf{91.69} &	84.09 &	85.17 &	74.60 &	\textbf{93.47} &	91.04 &	96.0 & \textbf{84.27}\\
 \midrule
 Xent-200 &  \textbf{89.36} &	97.96 &	86.49 &	\textbf{76.50} &	\textbf{64.36}  & 90.01 & 84.22 & \textbf{86.27} & \textbf{76.76} & \textbf{93.48} & 93.84 & \textbf{97.20} &  \textbf{85.77}\\
  SupCon-200 &  88.62 & \textbf{98.28} & \textbf{87.28} & 76.26 & 60.46 & \textbf{91.78} & \textbf{88.68} & 85.18 & 74.26 & 93.12 & \textbf{94.91} & 96.97 & \textbf{85.67}\\
  \bottomrule
\end{tabular}

\vspace{2mm}
\caption{Transfer learning results.  Numbers are mAP for VOC2007 \cite{pascal-voc-2007}; mean-per-class accuracy for Aircraft, Pets, Caltech, and Flowers; and top-1 accuracy for all other datasets.} %
\vspace{-22pt}
\label{table:transfer}
\end{table}

\subsection{Training Details}
The SupCon loss was trained for $700$ epochs during pretraining for ResNet-200 and $350$ epochs for smaller models. Fig. \ref{fig:hparam_stability}(c) shows accuracy as a function of SupCon training epochs for a ResNet50, demonstrating that even 200 epochs is likely sufficient for most purposes.

An (optional) additional step of training a linear classifier is used to compute top-1 accuracy. This is not needed if the purpose is to use representations for transfer learning tasks or retrieval. The second stage needs as few as 10 epochs of additional training. Note that in practice the linear classifier can be trained jointly with the encoder and projection networks by blocking gradient propagation from the linear classifier back to the encoder, and achieve roughly the same results without requiring two-stage training. We chose not to do that here to help isolate the effects of the SupCon loss.

We trained our models with batch sizes of up to 6144, although batch sizes of 2048 suffice for most purposes for both SupCon and cross-entropy losses (as shown in Fig. \ref{fig:hparam_stability}(b)). We associate some of the performance increase with batch size to the effect on the gradient due to hard positives increasing with an increasing number of negatives (see the Supplementary for details). We report metrics for experiments with batch size $6144$ for ResNet-50 and batch size $4096$ for ResNet-200 (due to the larger network size, a smaller batch size is necessary). We observed that for a fixed batch size it was possible to train with SupCon using larger learning rates than what was required by cross-entropy to achieve similar performance.

All our results used a temperature of $\tau = 0.1$. Smaller temperature benefits training more than higher ones, but extremely low temperatures are harder to train due to numerical instability. Fig. \ref{fig:hparam_stability}(d) shows the effect of temperature on Top-1 performance of supervised contrastive learning. As we can see from Eq. \ref{eqn:gradient}, the gradient scales inversely with choice of temperature $\tau$; therefore we rescale the loss by $\tau$ during training for stability.

We experimented with standard optimizers such as LARS \cite{you2017large}, RMSProp \cite{hinton2012neural} and SGD with momentum \cite{ruder2016overview} in different permutations for the initial pre-training step and training of the dense layer. While SGD with momentum works best for training ResNets with cross-entropy, we get the best performance for SupCon on ImageNet by using LARS for pre-training and RMSProp to training the linear layer. For CIFAR10 and CIFAR100 SGD with momentum performed best. Additional results for combinations of optimizers are provided in the Supplementary. Reference code is released at \url{https://t.ly/supcon}. 
{
\bibliographystyle{ieee_fullname}
\bibliography{supcon.bib}
}

\clearpage

\onecolumn
\begin{center}
{\Large\bf{Supplementary}\\
}
\end{center}

\maketitle

\section{Training Setup}
\begin{figure}[h]  
 \includegraphics[width=\linewidth]{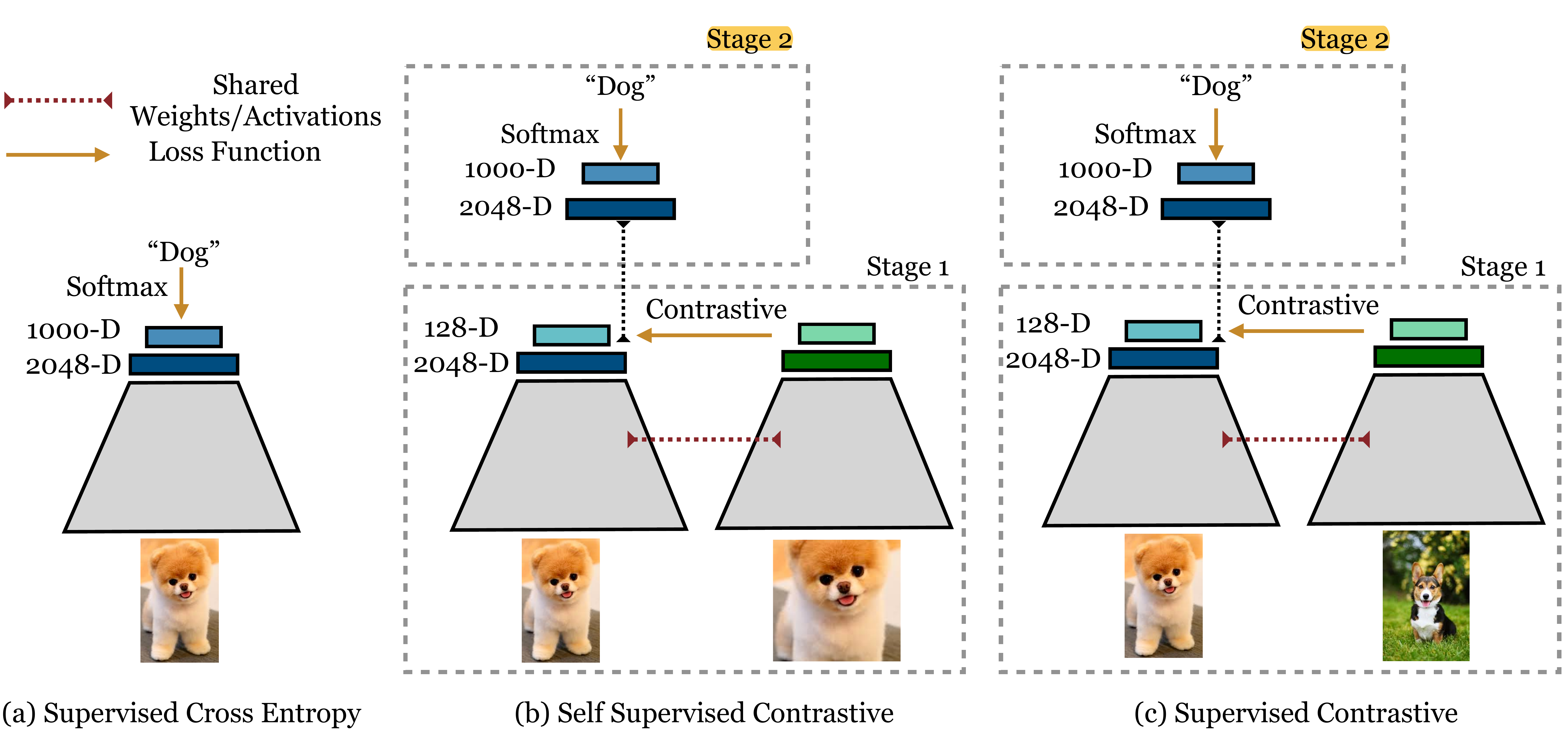}
  \caption{Cross entropy, self-supervised contrastive loss and supervised contrastive loss: The cross entropy loss (left) uses labels and a softmax loss to train a classifier; the self-supervised contrastive loss (middle) uses a contrastive loss and data augmentations to learn representations. The supervised contrastive loss (right) also learns representations using a contrastive loss, but uses label information to sample positives in addition to augmentations of the same image. Both contrastive methods can have an optional second stage which trains a model on top of the learned representations. }
  \label{fig:teaser2}
\end{figure}

In Fig. \ref{fig:teaser2}, we compare the training setup for the cross-entropy, self-supervised contrastive and supervised contrastive (SupCon) losses. Note that the number of parameters in the inference models always stays the same.  We also note that it is not necessary to train a linear classifier in the second stage, and previous works have used k-Nearest Neighbor classification \cite{wu2018unsupervised} or prototype classification to evaluate representations on classification tasks. The linear classifier can also be trained jointly with the encoder, as long as it doesn't propagate gradients back to the encoder. 

\section{Gradient Derivation}
\label{sec:gradient_derivation}
In Sec. 3 of the paper, we make the claim that the gradients of the two considered supervised contrastive losses, $\mathcal{L}_{out}^{sup}$ and $\mathcal{L}_{in}^{sup}$, with respect to a normalized projection network representation, $\boldsymbol{z}_i$, have a nearly identical mathematical form. In this section, we perform derivations to show this is true. It is sufficient to show that this claim is true for $\mathcal{L}_{out,i}^{sup}$ and $\mathcal{L}_{in,i}^{sup}$. For convenience, we reprint below the expressions for each.
\begin{equation}
  \mathcal{L}_{in,i}^{sup}
  =-\log\left\{\frac{1}{|P(i)|}\sum_{p\in P(i)}\frac{\text{exp}\left(\boldsymbol{z}_i\bigcdot\boldsymbol{z}_p/\tau\right)}{\sum\limits_{a\in A(i)}\text{exp}\left(\boldsymbol{z}_i\bigcdot\boldsymbol{z}_a/\tau\right)}\right\}
  \label{eqn:supp_bad_supervised_loss}
\end{equation}
\begin{equation}
  \mathcal{L}_{out,i}^{sup}
  =\frac{-1}{|P(i)|}\sum_{p\in P(i)}\log{\frac{\text{exp}\left(\boldsymbol{z}_i\bigcdot\boldsymbol{z}_p/\tau\right)}{\sum\limits_{a\in A(i)}\text{exp}\left(\boldsymbol{z}_i\bigcdot\boldsymbol{z}_a/\tau\right)}}
  \label{eqn:supp_supervised_loss}
\end{equation}

We start by deriving the gradient of $\mathcal{L}_{in}^{sup}$ (Eq. \ref{eqn:supp_bad_supervised_loss}):
\begin{align}
  \frac{\partial\mathcal{L}_{in}^{sup}}{\partial\boldsymbol{z}_i} & = -\frac{\partial}{\partial\boldsymbol{z}_i}\log\left\{\frac{1}{|P(i)|}\sum_{p\in P(i)}\frac{\text{exp}\left(\boldsymbol{z}_i\bigcdot\boldsymbol{z}_p/\tau\right)}{\sum\limits_{a\in A(i)}\text{exp}\left(\boldsymbol{z}_i\bigcdot\boldsymbol{z}_a/\tau\right)}\right\}\nonumber\\
  & = \frac{\partial}{\partial\boldsymbol{z}_i}\log\sum\limits_{a\in A(i)}\text{exp}\left(\boldsymbol{z}_i\bigcdot\boldsymbol{z}_a/\tau\right)-\frac{\partial}{\partial\boldsymbol{z}_i}\log\sum\limits_{p\in P(i)}\text{exp}\left(\boldsymbol{z}_i\bigcdot\boldsymbol{z}_p/\tau\right)\nonumber\\
  & = \frac{1}{\tau}\frac{\sum\limits_{a\in A(i)}\boldsymbol{z}_a\text{exp}\left(\boldsymbol{z}_i\bigcdot\boldsymbol{z}_a/\tau\right)}{\sum\limits_{a\in A(i)}\text{exp}\left(\boldsymbol{z}_i\bigcdot\boldsymbol{z}_a/\tau\right)}-\frac{1}{\tau}\frac{\sum\limits_{p\in P(i)}\boldsymbol{z}_p\text{exp}\left(\boldsymbol{z}_i\bigcdot\boldsymbol{z}_p/\tau\right)}{\sum\limits_{p\in P(i)}\text{exp}\left(\boldsymbol{z}_i\bigcdot\boldsymbol{z}_p/\tau\right)}\nonumber\\
  & = \frac{1}{\tau}\frac{\sum\limits_{p\in P(i)}\boldsymbol{z}_p\text{exp}\left(\boldsymbol{z}_i\bigcdot\boldsymbol{z}_p/\tau\right)+\sum\limits_{n\in N(i)}\boldsymbol{z}_n\text{exp}\left(\boldsymbol{z}_i\bigcdot\boldsymbol{z}_n/\tau\right)}{\sum\limits_{a\in A(i)}\text{exp}\left(\boldsymbol{z}_i\bigcdot\boldsymbol{z}_a/\tau\right)}-\frac{1}{\tau}\frac{\sum\limits_{p\in P(i)}\boldsymbol{z}_p\text{exp}\left(\boldsymbol{z}_i\bigcdot\boldsymbol{z}_p/\tau\right)}{\sum\limits_{p\in P(i)}\text{exp}\left(\boldsymbol{z}_i\bigcdot\boldsymbol{z}_p/\tau\right)}\nonumber\\
  & = \frac{1}{\tau}\left\{\sum_{p\in P(i)}\boldsymbol{z}_p(P_{ip}-X_{ip}^{in})+\sum_{n\in N(i)}\boldsymbol{z}_nP_{in}\right\}
  \label{eqn:supp_bad_supervised_loss_gradient}
\end{align}
where we have defined:
\begin{align}
  P_{ip}\equiv\frac{\text{exp}\left(\boldsymbol{z}_i\bigcdot\boldsymbol{z}_p/\tau\right)}{\sum_{a\in A(i)}\text{exp}\left(\boldsymbol{z}_i\bigcdot\boldsymbol{z}_a/\tau\right)}
  \label{eqn:supp_pip} \\
  X_{ip}^{in}\equiv\frac{\text{exp}\left(\boldsymbol{z}_i\bigcdot\boldsymbol{z}_p/\tau\right)}{\sum\limits_{p'\in P(i)}\text{exp}\left(\boldsymbol{z}_i\bigcdot\boldsymbol{z}_{p'}/\tau\right)}
\end{align}
Though similar in structure, $P_{ip}$ and $X_{ip}^{in}$ are fundamentally different: $P_{ip}$ is the likelihood for $\boldsymbol{z}_p$ with respect to all positives and negatives, while $X_{ip}^{in}$ is that but with respect to only the positives. $P_{in}$ is analogous to $P_{ip}$ but defines the likelihood of $\boldsymbol{z}_n$. In particular, $P_{ip}\le X_{ip}^{in}$. We now derive the gradient of Eq. \ref{eqn:supp_supervised_loss}:
\begin{align}
  \frac{\partial\mathcal{L}_{out}^{sup}}{\partial\boldsymbol{z}_i} & = \frac{-1}{|P(i)|}\sum_{p\in P(i)}\frac{\partial}{\partial\boldsymbol{z}_i}\left\{\frac{\boldsymbol{z}_i\bigcdot\boldsymbol{z}_p}{\tau}-\log\sum\limits_{a\in A(i)}\text{exp}\left(\boldsymbol{z}_i\bigcdot\boldsymbol{z}_a/\tau\right)\right\}\nonumber\\
  & = \frac{-1}{\tau|P(i)|}\sum_{p\in P(i)}\left\{\boldsymbol{z}_p-\frac{\sum\limits_{a\in A(i)}\boldsymbol{z}_a\text{exp}\left(\boldsymbol{z}_i\bigcdot\boldsymbol{z}_a/\tau\right)}{\sum\limits_{a\in A(i)}\text{exp}\left(\boldsymbol{z}_i\bigcdot\boldsymbol{z}_a/\tau\right)}\right\}\nonumber\\
  & = \frac{-1}{\tau|P(i)|}\sum_{p\in P(i)}\left\{\boldsymbol{z}_p-\sum\limits_{p'\in P(i)}\boldsymbol{z}_{p'}P_{ip'}-\sum\limits_{n\in N(i)}\boldsymbol{z}_nP_{in}\right\}\nonumber\\
  & = \frac{-1}{\tau|P(i)|}\left\{\sum_{p\in P(i)}\boldsymbol{z}_p-\sum_{p\in P(i)}\sum\limits_{p'\in P(i)}\boldsymbol{z}_{p'}P_{ip'}-\sum_{p\in P(i)}\sum\limits_{n\in N(i)}\boldsymbol{z}_nP_{in}\right\}\nonumber\\
  & = \frac{-1}{\tau|P(i)|}\left\{\sum_{p\in P(i)}\boldsymbol{z}_p-\sum_{p'\in P(i)}\sum\limits_{p\in P(i)}\boldsymbol{z}_{p'}P_{ip'}-\sum_{n\in N(i)}\sum\limits_{p\in P(i)}\boldsymbol{z}_nP_{in}\right\}\nonumber\\
  & = \frac{-1}{\tau|P(i)|}\left\{\sum_{p\in P(i)}\boldsymbol{z}_p-\sum_{p'\in P(i)}|P(i)|\boldsymbol{z}_{p'}P_{ip'}-\sum_{n\in N(i)}|P(i)|\boldsymbol{z}_nP_{in}\right\}\nonumber\\
  & = \frac{-1}{\tau|P(i)|}\left\{\sum_{p\in P(i)}\boldsymbol{z}_p-\sum_{p\in P(i)}|P(i)|\boldsymbol{z}_pP_{ip}-\sum_{n\in N(i)}|P(i)|\boldsymbol{z}_nP_{in}\right\}\nonumber\\
  & = \frac{1}{\tau}\left\{\sum_{p\in P(i)}\boldsymbol{z}_p(P_{ip}-X_{ip}^{out})+\sum_{n\in N(i)}\boldsymbol{z}_nP_{in}\right\}
  \label{eqn:supp_supervised_loss_gradient}
\end{align}
where we have defined:
\begin{equation}
  X_{ip}^{out}\equiv\frac{1}{|P(i)|}
\end{equation}
Thus, both gradients (Eqs. \ref{eqn:supp_bad_supervised_loss_gradient} and \ref{eqn:supp_supervised_loss_gradient}) have a very similar form and can be written collectively as:
\begin{equation}
  \label{eqn:supp_gradient}
  \frac{\partial\mathcal{L}_i^{sup}}{\partial\boldsymbol{z}_i} = \frac{1}{\tau}\left\{
  \sum_{p\in P(i)}\boldsymbol{z}_p(P_{ip}-X_{ip})+
  \sum_{n\in N(i)}\boldsymbol{z}_nP_{in}
  \right\}
\end{equation}
where:
\begin{equation}
  X_{ip}\equiv\left\{
  \begin{matrix}
  \frac{\text{exp}\left(\boldsymbol{z}_i\bigcdot\boldsymbol{z}_p/\tau\right)}{\sum\limits_{p'\in P(i)}\text{exp}\left(\boldsymbol{z}_i\bigcdot\boldsymbol{z}_{p'}/\tau\right)} & , & \text{if }\mathcal{L}_i^{sup}=\mathcal{L}_{in,i}^{sup}\\
  \frac{1}{|P(i)|} & , & \text{if }\mathcal{L}_i^{sup}=\mathcal{L}_{out,i}^{sup}
  \end{matrix}\right.
\end{equation}

This corresponds to Eq. 4 and subsequent analysis in the paper.

\section{Intrinsic Hard Positive and Negative Mining Properties}
\label{sec:hard_mining_properties}
The contrastive loss is structured so that gradients with respect to the \emph{unnormalized} projection network representations provide an intrinsic mechanism for hard positive/negative mining during training. For losses such as the triplet loss or max-margin, hard mining is known to be crucial to their performance. For contrastive loss, we show analytically that hard mining is intrinsic and thus removes the need for complicated hard mining algorithms.

As shown in Sec. \ref{sec:gradient_derivation}, the gradients of both $\mathcal{L}_{out}^{sup}$ and $\mathcal{L}_{in}^{sup}$ are given by Eq. \ref{eqn:supp_supervised_loss_gradient}. Additionally, note that the self-supervised contrastive loss, $\mathcal{L}_i^{self}$, is a special case of either of the two supervised contrastive losses (when $P(i)=j(i)$). So by showing that Eq. \ref{eqn:supp_supervised_loss_gradient} has structure that provides hard positive/negative mining, it will be shown to be true for all three contrastive losses (self-supervised and both supervised versions).

The projection network applies a normalization to its outputs\footnote{Note that when the normalization is combined with an inner product (as we do here), this is equivalent to cosine similarity. Some contrastive learning approaches \cite{chen2020simple} use a cosine similarity explicitly in their loss formulation. We decouple the normalization here to highlight the benefits it provides.%
}. We shall let $\boldsymbol{w}_i$ denote the projection network output \emph{prior} to normalization, i.e., $\boldsymbol{z}_i=\boldsymbol{w}_i/\Vert\boldsymbol{w}_i\Vert$. As we show below, normalizing the representations provides structure (when combined with Eq. \ref{eqn:supp_supervised_loss_gradient}) to the gradient enables the learning to focus on hard positives and negatives. The gradient of the supervised loss with respect to $\boldsymbol{w}_i$ is related to that with respect to $\boldsymbol{z}_i$ via the chain rule:
\begin{equation}
  \frac{\partial\mathcal{L}_i^{sup}(\boldsymbol{z}_i)}{\partial\boldsymbol{w}_i}=\frac{\partial\boldsymbol{z}_i}{\partial\boldsymbol{w}_i}\frac{\partial\mathcal{L}_i^{sup}(\boldsymbol{z}_i)}{\partial\boldsymbol{z}_i} \label{eqn:supp_gradient_chain_rule}
\end{equation}
where:
\begin{align}
  \frac{\partial\boldsymbol{z}_i}{\partial\boldsymbol{w}_i} & = \frac{\partial}{\partial\boldsymbol{w}_i}\left(\frac{\boldsymbol{w}_i}{\Vert\boldsymbol{w}_i\Vert}\right) \nonumber \\
  & = \frac{1}{\Vert\boldsymbol{w}_i\Vert}\text{I}-\boldsymbol{w}_i\left(\frac{\partial\left(1/\Vert\boldsymbol{w}_i\Vert\right)}{\partial\boldsymbol{w}_i}\right)^T \nonumber \\
  & = \frac{1}{\Vert\boldsymbol{w}_i\Vert}\left(\text{I}-\frac{\boldsymbol{w}_i\boldsymbol{w}_i^T}{\Vert\boldsymbol{w}_i\Vert^2}\right) \nonumber \\
  & = \frac{1}{\Vert\boldsymbol{w}_i\Vert}\left(\text{I}-\boldsymbol{z}_i\boldsymbol{z}_i^T\right)
  \label{eqn:supp_dz_dx}
\end{align}
Combining Eqs. \ref{eqn:supp_supervised_loss_gradient} and \ref{eqn:supp_dz_dx} thus gives:
\begin{align}
  \frac{\partial\mathcal{L}_i^{sup}}{\partial\boldsymbol{w}_i} & = \frac{1}{\tau\Vert\boldsymbol{w}_i\Vert}\left(\text{I}-\boldsymbol{z}_i\boldsymbol{z}_i^T\right)\left\{\sum_{p\in P(i)}\boldsymbol{z}_p(P_{ip}-X_{ip})+\sum_{n\in N(i)}\boldsymbol{z}_nP_{in}\right\}\nonumber\\
  & = \frac{1}{\tau\Vert\boldsymbol{w}_i\Vert}\left\{\sum_{p\in P(i)}(\boldsymbol{z}_p-(\boldsymbol{z}_i\bigcdot\boldsymbol{z}_p)\boldsymbol{z}_i)(P_{ip}-X_{ip})+\sum_{n\in N(i)}(\boldsymbol{z}_n-(\boldsymbol{z}_i\bigcdot\boldsymbol{z}_n)\boldsymbol{z}_i)P_{in}\right\}\nonumber\\
  & = \left.\frac{\partial\mathcal{L}_i^{sup}}{\partial\boldsymbol{w}_i}\right\vert_\text{P(i)}+\left.\frac{\partial\mathcal{L}_i^{sup}}{\partial\boldsymbol{w}_i}\right\vert_\text{N(i)}
\end{align}
where:
\begin{align}
  \left.\frac{\partial\mathcal{L}_i^{sup}}{\partial\boldsymbol{w}_i}\right\vert_\text{P(i)} & =
  \frac{1}{\tau\Vert\boldsymbol{w}_i\Vert}\sum_{p\in P(i)}(\boldsymbol{z}_p-(\boldsymbol{z}_i\bigcdot\boldsymbol{z}_p)\boldsymbol{z}_i)(P_{ip}-X_{ip})\label{eqn:supp_loss_gradient_pos}\\
  \left.\frac{\partial\mathcal{L}_i^{sup}}{\partial\boldsymbol{w}_i}\right\vert_\text{N(i)} & =
  \frac{1}{\tau\Vert\boldsymbol{w}_i\Vert}\sum_{n\in N(i)}(\boldsymbol{z}_n-(\boldsymbol{z}_i\bigcdot\boldsymbol{z}_n)\boldsymbol{z}_i)P_{in}\label{eqn:supp_loss_gradient_neg}
\end{align}

We now show that easy positives and negatives have small gradient contributions while hard positives and negatives have large ones. For an easy positive (i.e., one against which contrasting the anchor only \emph{weakly} benefits the encoder), $\boldsymbol{z}_i\bigcdot\boldsymbol{z}_p\approx 1$. Thus (see Eq. \ref{eqn:supp_loss_gradient_pos}):
\begin{equation}
  \left\Vert(\boldsymbol{z}_p-(\boldsymbol{z}_i\bigcdot\boldsymbol{z}_p)\boldsymbol{z}_i\right\Vert=\sqrt{1-(\boldsymbol{z}_i\bigcdot\boldsymbol{z}_p)^2}\approx 0
\end{equation}
However, for a hard positive (i.e., one against which contrasting the anchor \emph{greatly} benefits the encoder), $\boldsymbol{z}_i\bigcdot\boldsymbol{z}_p\approx 0$, so:
\begin{equation}
  \left\Vert(\boldsymbol{z}_p-(\boldsymbol{z}_i\bigcdot\boldsymbol{z}_p)\boldsymbol{z}_i\right\Vert=\sqrt{1-(\boldsymbol{z}_i\bigcdot\boldsymbol{z}_p)^2}\approx 1
\end{equation}
Thus, for the gradient of $\mathcal{L}_{in}^{sup}$ (where $X_{ip}=X_{ip}^{in}$):
\begin{align}
  & \left\Vert(\boldsymbol{z}_p-(\boldsymbol{z}_i\bigcdot\boldsymbol{z}_p)\boldsymbol{z}_i\right\Vert|P_{ip}-X_{ip}^{in}|\nonumber\\
  & \approx |P_{ip}-X_{ip}^{in}|\nonumber\\
  & = \left|\frac{1}{\sum\limits_{p'\in P(i)}\text{exp}\left(\boldsymbol{z}_i\bigcdot\boldsymbol{z}_{p'}/\tau\right)+\sum\limits_{n\in N(i)}\text{exp}\left(\boldsymbol{z}_i\bigcdot\boldsymbol{z}_n/\tau\right)}-\frac{1}{\sum\limits_{p'\in P(i)}\text{exp}\left(\boldsymbol{z}_i\bigcdot\boldsymbol{z}_{p'}/\tau\right)}\right|\nonumber\\
  & \propto \sum_{n\in N(i)}\text{exp}\left(\boldsymbol{z}_i\bigcdot\boldsymbol{z}_n/\tau\right)
  \label{eqn:supp_bad_supcon_factor}
\end{align}
For the gradient of $\mathcal{L}_{out}^{sup}$ (where $X_{ip}=X_{ip}^{out}$)
\begin{align}
  & \left\Vert(\boldsymbol{z}_p-(\boldsymbol{z}_i\bigcdot\boldsymbol{z}_p)\boldsymbol{z}_i\right\Vert|P_{ip}-X_{ip}^{out}|\nonumber\\
  & \approx |P_{ip}-X_{ip}^{out}|\nonumber\\
  & = \left|\frac{1}{\sum\limits_{a\in A(i)}\text{exp}\left(\boldsymbol{z}_i\bigcdot\boldsymbol{z}_a/\tau\right)}-\frac{1}{|P(i)|}\right|\nonumber\\
  & = \left|\frac{1}{\sum\limits_{p'\in P(i)}\text{exp}\left(\boldsymbol{z}_i\bigcdot\boldsymbol{z}_{p'}/\tau\right)+\sum\limits_{n\in N(i)}\text{exp}\left(\boldsymbol{z}_i\bigcdot\boldsymbol{z}_n/\tau\right)}-\frac{1}{|P(i)|}\right|\nonumber\\
  & \propto \sum_{n\in N(i)}\text{exp}\left(\boldsymbol{z}_i\bigcdot\boldsymbol{z}_n/\tau\right)+\sum_{p'\in P(i)}\text{exp}\left(\boldsymbol{z}_i\bigcdot\boldsymbol{z}_{p'}/\tau\right)-|P(i)|
  \label{eqn:supp_supcon_factor}
\end{align}
where $\sum_{n\in N(i)}\text{exp}(\boldsymbol{z}_i\bigcdot\boldsymbol{z}_n/\tau)\ge0$ (assuming $\boldsymbol{z}_i\bigcdot\boldsymbol{z}_n\le0$) and $\sum_{p'\in P(i)}\text{exp}(\boldsymbol{z}_i\bigcdot\boldsymbol{z}_{p'}/\tau)-|P(i)|\ge0$ (assuming $\boldsymbol{z}_i\bigcdot\boldsymbol{z}_{p'}\ge0$). We thus see that for either $\mathcal{L}_{out}^{sup}$ and $\mathcal{L}_{in}^{sup}$ the gradient response to a hard positive in any individual training step can be made larger by increasing the number of negatives. 
Additionally, for $\mathcal{L}_{out}^{sup}$, it can also be made larger by increasing the number of positives. 

Thus, for weak positives (since $\boldsymbol{z}_i\bigcdot\boldsymbol{z}_p\approx 1$) the contribution to the gradient is small while for hard positives the contribution is large (since $\boldsymbol{z}_i\bigcdot\boldsymbol{z}_p \approx 0$). Similarly, analysing Eq. \ref{eqn:supp_loss_gradient_neg} for weak negatives ($\boldsymbol{z}_i\bigcdot\boldsymbol{z}_n\approx -1$) vs hard negatives ($\boldsymbol{z}_i\bigcdot\boldsymbol{z}_n\approx 0$) we conclude that the gradient contribution is large for hard negatives and small for weak negatives.

In addition to an increased number of positives/negatives helping in general, we also note that as we increase the batch size, we also increase the probability of choosing individual \emph{hard} positives/negatives. Since hard positives/negatives lead to a larger gradient contribution, we see that a larger batch has multiple high impact effects to allow obtaining better performance, as we observe empirically in the main paper. %

Additionally, it should be noted that the ability of contrastive losses to perform intrinsic hard positive/negative data mining comes about only if a normalization layer is added to the end of the projection network, thereby justifying the use of a normalization in the network. Ours is the first paper to show analytically this property of contrastive losses, even though normalization has been empirically found to be useful in self-supervised contrastive learning.

\section{Triplet Loss Derivation from Contrastive Loss}
\label{sec:triplet_loss_derivation}

In this section, we show that the triplet loss \cite{weinberger2009distance} is a special case of the contrastive loss when the number of positives and negatives are each one. Assuming the representation of the anchor ($i$) and the positive ($p$) are more aligned than that of the anchor and negative ($n$) (i.e., $\boldsymbol{z}_i\bigcdot\boldsymbol{z}_p\gg\boldsymbol{z}_i\bigcdot\boldsymbol{z}_n$), we have:
\begin{eqnarray}
  \mathcal{L}^{self} & = & -\text{log}\frac{\text{exp}\left(\boldsymbol{z}_a\bigcdot\boldsymbol{z}_p/\tau\right)}{\text{exp}\left(\boldsymbol{z}_a\bigcdot\boldsymbol{z}_p/\tau\right)+\text{exp}\left(\boldsymbol{z}_a\bigcdot\boldsymbol{z}_n/\tau\right)} \nonumber \\
  & = & \text{log}\left(1+\text{exp}\left(\left(\boldsymbol{z}_a\bigcdot\boldsymbol{z}_n-\boldsymbol{z}_a\bigcdot\boldsymbol{z}_p\right)/\tau\right)\right) \nonumber \\
  & \approx & \text{exp}\left(\left(\boldsymbol{z}_a\bigcdot\boldsymbol{z}_n-\boldsymbol{z}_a\bigcdot\boldsymbol{z}_p\right)/\tau\right) ~~~~\text{(Taylor expansion of $\log$)} \nonumber \\
  & \approx & 1+\frac{1}{\tau}\cdot\left(\boldsymbol{z}_a\bigcdot\boldsymbol{z}_n-\boldsymbol{z}_a\bigcdot\boldsymbol{z}_p\right) \nonumber \\
  & = & 1-\frac{1}{2\tau}\cdot\left(\left\Vert\boldsymbol{z}_a-\boldsymbol{z}_n\right\Vert^2-\left\Vert\boldsymbol{z}_a-\boldsymbol{z}_p\right\Vert^2\right) \nonumber \\
  & \propto & \left\Vert\boldsymbol{z}_a-\boldsymbol{z}_p\right\Vert^2-\left\Vert\boldsymbol{z}_a-\boldsymbol{z}_n\right\Vert^2+2\tau \nonumber
\end{eqnarray}
which has the same form as a triplet loss with margin $\alpha=2\tau$. This result is consistent with empirical results \cite{chen2020simple} which show that contrastive loss performs better in general than triplet loss on representation tasks. Additionally, whereas triplet loss in practice requires computationally expensive hard negative mining (e.g., \cite{schroff2015facenet}), the discussion in Sec. \ref{sec:hard_mining_properties} shows that the gradients of the supervised contrastive loss naturally impose a measure of hard negative reinforcement during training. This comes at the cost of requiring large batch sizes to include many positives and negatives.

\section{Supervised Contrastive Loss Hierarchy}

The SupCon loss subsumes multiple other commonly used losses as special cases of itself. It is insightful to study which additional restrictions need to be imposed on it to change its form into that of each of these other losses.

For convenience, we reprint the form of the SupCon loss.
\begin{equation}
  \mathcal{L}^{sup}
  =\sum_{i\in I}\frac{-1}{|P(i)|}\sum_{p\in P(i)}\log{\frac{\text{exp}\left(\boldsymbol{z}_i\bigcdot\boldsymbol{z}_p/\tau\right)}{\sum\limits_{a\in A(i)}\text{exp}\left(\boldsymbol{z}_i\bigcdot\boldsymbol{z}_a/\tau\right)}}
  \label{eqn:supp_supervised_loss}
\end{equation}

Here, $P(i)$ is the set of all positives in the multiviewed batch corresponding to the anchor $i$. For SupCon, positives can come from two disjoint categories:
\begin{itemize}
  \item Views of the \emph{same} sample image which generated the anchor image.
  \item Views of a sample image \emph{different} from that which generated the anchor image but having the same label as that of the anchor.
\end{itemize}

The loss for self-supervised contrastive learning (Eq. 1 in the paper) is a special case of SupCon when $P(i)$ is restricted to contain only a view of the \emph{same} source image as that of the anchor (i.e., the first category above). In this case, $P(i)=j(i)$, where $j(i)$ is the index of view, and Eq. \ref{eqn:supp_supervised_loss} readily takes on the self-supervised contrastive loss form.
\begin{equation}
  \left.\mathcal{L}^{sup}\right|_{P(i)=j(i)}
  =\mathcal{L}^{self}
  =-\sum_{i\in I}\log{
  \frac{\text{exp}\left(\boldsymbol{z}_i\bigcdot\boldsymbol{z}_{j(i)}/\tau\right)}{\sum\limits_{a\in A(i)}\text{exp}\left(\boldsymbol{z}_i\bigcdot\boldsymbol{z}_a/\tau\right)}
  }
  \label{eqn:supp_self_loss}
\end{equation}

A second commonly referenced loss subsumed by SupCon is the N-Pairs loss \cite{sohn2016improved}. This loss, while functionally similar to Eq. \ref{eqn:supp_self_loss}, differs from it by requiring that the positive be generated from a sample image \emph{different} from that which generated the anchor but which has the same label as the anchor (i.e., the second category above). There is also no notion of temperature in the original N-Pairs loss, though it could be easily generalized to include it. Letting $k(i)$ denote the positive originating from a different sample image than that which generated the anchor $i$, the N-Pairs loss has the following form:
\begin{equation}
  \left.\mathcal{L}^{sup}\right|_{P(i)=k(i),\tau=1}
  =\mathcal{L}^{n{\text -}pairs}
  =-\sum_{i\in I}\log{
  \frac{\text{exp}\left(\boldsymbol{z}_i\bigcdot\boldsymbol{z}_{k(i)}\right)}{\sum\limits_{a\in A(i)}\text{exp}\left(\boldsymbol{z}_i\bigcdot\boldsymbol{z}_a\right)}
  }
  \label{eqn:supp_npairs_loss}
\end{equation}
It is interesting to see how these constraints affect performance. For a batch size of 6144, a ResNet-50 encoder trained on ImageNet with N-Pairs loss achieves an ImageNet Top-1 classification accuracy of 57.4\% while an identical setup trained with the SupCon loss achieves 78.7\%.

Finally, as discussed in Sec. \ref{sec:triplet_loss_derivation}, triplet loss is a special case of the SupCon loss (as well as that of the self-supervised and N-Pairs losses) when the number of positives and negatives are restricted to both be one.

\section{Effect of Temperature in Loss Function}

Similar to previous work \cite{chen2020simple,tian2019contrastive}, we find that the temperature $\tau$ used in the loss function has an important role to play in supervised contrastive learning and that the model trained with the optimal temperature can improve performance by nearly 3\%. Two competing effects that changing the temperature has on training the model are:
\begin{enumerate}
    \item {\bf Smoothness:} The distances in the representation space used for training the model have gradients with smaller norm ($||\nabla \mathcal{L}|| \propto \frac{1}{\tau}$); see Section \ref{sec:gradient_derivation}. Smaller magnitude gradients make the optimization problem simpler by allowing for larger learning rates. In Section 3.3 of the paper, it is shown that in the case of a single positive and negative, the contrastive loss is equivalent to a triplet loss with margin $\propto\tau$. Therefore, in these cases, a larger temperature makes the optimization easier, and classes more separated.
    
    \item {\bf Hard positives/negatives:} On the other hand, as shown in Sec \ref{sec:hard_mining_properties}, the supervised contrastive loss has structure that cause hard positives/negatives to improve performance. Additionally, hard negatives have been shown to improve classification accuracy when models are trained with the triplet loss \cite{schroff2015facenet}. Low temperatures are equivalent to optimizing for hard positives/negatives: for a given batch of samples and a specific anchor, lowering the temperature relatively increases the value of $P_{ik}$ (see Eq. \ref{eqn:supp_pip}) for samples which have larger inner product with the anchor, and reduces it for samples which have smaller inner product.

\end{enumerate}

We found empirically that a temperature of $0.1$ was optimal for top-1 accuracy on ResNet-50; results on various temperatures are shown in Fig. 4 of the main paper. We use the same temperature for all experiments on ResNet-200.

\section{Effect of Number of Positives} 
\begin{wraptable}{r}{0.5\textwidth}
    \small
    \vspace{-15mm}
    \centering
    \begin{tabular}{cccccc}
    \toprule
     1 \cite{chen2020simple} & 3 & 5 & 7 & 9 & No cap (13) \\\midrule
    69.3 & 76.6 & 78.0 & 78.4 & 78.3 & 78.5 \\\bottomrule
    \end{tabular}    %
    \caption{Comparison of Top-1 accuracy variability as a function of the maximum number of positives $|P(i)|$  varies from 1 to no cap . Adding more positives benefits the final Top-1 accuracy. Note that with 1 positive, this is equivalent to the self-supervised approach of \cite{chen2020simple} where the positive is an augmented version of the \emph{same sample}. }
    \label{tab:pos_ablation}
\end{wraptable}

We run ablations to test the effect of the number of positives. Specifically, we take at most $k$ positives for each sample, and also remove them from the denominator of the loss function so they are not considered as a negative. We train with a batch size of 6144, so without this capping there are 13 positives in expectation(6 positives, each with 2 augmentatioins, plus other augmentation of anchor image). We train for 350 epochs. Table \ref{tab:pos_ablation} shows the steady benefit of adding more positives for a ResNet-50 model trained on ImageNet with supervised contrastive loss.  Note that for each anchor, the number of positives always contains one positive which is the same sample but with a different data augmentation; and the remainder of the positives are different samples from the same class. Under this definition, self-supervised learning is considered as having $1$ positive.

\section{Robustness}
\begin{figure*}  
    \centering
    \includegraphics[width=0.45\linewidth]{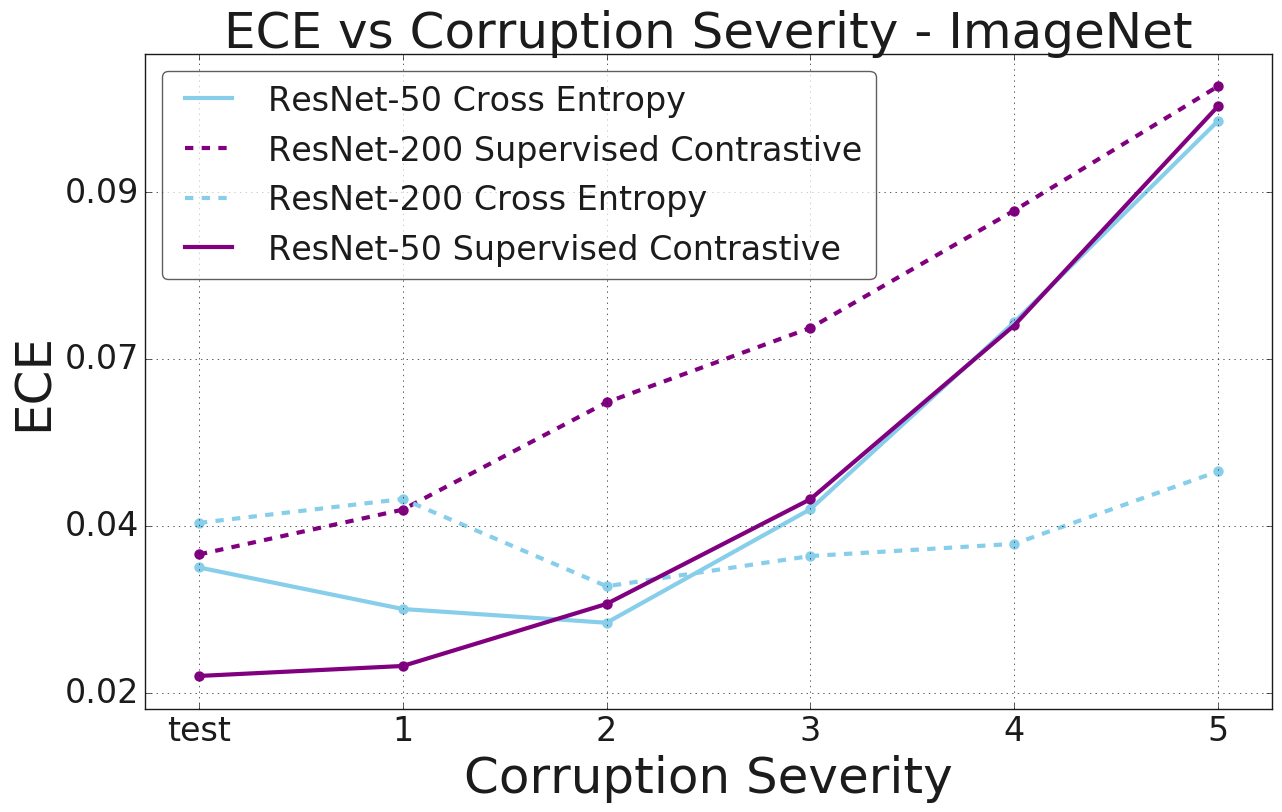}
    \caption{Expected Calibration Error and mean top-1 accuracy at different corruption severities on ImageNet-C, on the ResNet-50 architecture (top) and ResNet-200 architecture (bottom). The contrastive loss maintains a higher accuracy over the range of corruption severities, and does not suffer from increasing calibration error, unlike the cross entropy loss.}
  \label{fig:robustness}
\end{figure*}

Along with measuring the mean Corruption Error (mCE) and mean relative Corruption Error \cite{hendrycks2019benchmarking} on the ImageNet-C dataset (see paper, Section 4.2 and Figure 3), we also measure the Expected Calibration Error and the mean accuracy of our models on different corruption severity levels. Table \ref{tab:severity} demonstrates how performance and calibration degrades as the data shifts farther from the training distribution and becomes harder to classify. Figure \ref{fig:robustness} shows how the calibration error of the model increases as the level of corruption severity increases as measured by performance on ImageNet-C \cite{hendrycks2019benchmarking}.

\begin{table}[ht]
    \centering
    \begin{tabular}{p{0.17\textwidth}llccccc}\toprule
        \multicolumn{2}{c}{Model}&
        Test & {1} & {2} & {3} & {4} & {5} \\\midrule
        Loss& Architecture & \multicolumn{6}{c}{ECE}\\\midrule
        \multirow{2}{0.17\textwidth}{Cross Entropy} & ResNet-50  & 0.039 & 0.033 & 0.032 & 0.047 & 0.072 & 0.098\\
        & ResNet-200  & 0.045 & 0.048 & 0.036 & 0.040 & 0.042 & 0.052 \\ 
        \midrule
        \multirow{2}{0.17\textwidth}{Supervised  Contrastive} & ResNet-50  & 0.024 & 0.026 & 0.034 & 0.048 & 0.071 & 0.100 \\ 
        & ResNet-200  & 0.041 & 0.047 & 0.061 & 0.071 & 0.086 & 0.103 \\\midrule
        & & \multicolumn{6}{c}{Top-1 Accuracy}\\\midrule
        \multirow{2}{0.17\textwidth}{Cross Entropy} & ResNet-50  & 78.24 & 65.06 & 54.96 & 47.64 & 35.93 & 25.38 \\ 
        & ResNet-200  & 80.81 & 72.89 & 65.28 & 60.55 & 52.00 & 43.11\\
        \midrule
        \multirow{2}{0.17\textwidth}{Supervised Contrastive} & ResNet-50  & 78.81 & 65.39 & 55.55 & 48.64 & 37.27 & 26.92 \\
         & ResNet-200  & 81.38 & 73.29 & 66.16 & 61.80 & 54.01 & 45.71 \\ \bottomrule
    \end{tabular}
    \vspace{2mm}
    \caption{{\bf{Top}}: Average Expected Calibration Error (ECE) over all the corruptions in ImageNet-C \cite{hendrycks2019benchmarking} for a given level of severity (lower is better); {\bf{Bottom}}: Average Top-1 Accuracy over all the corruptions for a given level of severity (higher is better).}
    \label{tab:severity}
\end{table}

\section{Two stage training on Cross Entropy}
To ablate the effect of representation learning and have a two stage evaluation process we also compared against using models trained with cross-entropy loss for representation learning. We do this by first training the model with cross entropy and then re-initializing the final layer of the network randomly. In this second stage of training we again train with cross entropy but keep the weights of the network fixed. Table \ref{table:xenthead} shows that the representations learnt by cross-entropy for a ResNet-50 network are not robust and just the re-initialization of the last layer leads to large drop in accuracy and a mixed result on robustness compared to a single-stage cross-entropy training. Hence both methods of training cross-entropy are inferior to supervised contrastive loss.

\begin{table}[ht]
    \centering
    \begin{tabular}{cccc}\toprule
         & Accuracy & mCE & rel. mCE  \\\midrule
        Supervised Contrastive & {\bf 78.7} & {\bf 67.2} & 94.6 \\
        Cross Entropy (1 stage) & 77.1 & 68.4 & 103.7  \\
        Cross Entropy (2 stage) & 73.7 & 73.3 & {\bf 92.9} \\\bottomrule
    \end{tabular}
    \caption{Comparison between representations learnt using Supervised Contrastive and representations learnt using Cross Entropy loss with either 1 stage of training or 2 stages (representation learning followed by linear classifier).}
    \label{table:xenthead}
\end{table}

\section{Training Details}
In this section we present results for various ablation experiments, disentangling the effects of (a) Optimizer and (b) Data Augmentation on downstream performance.
\subsection{Optimizer}
We experiment with various optimizers for the contrastive learning and training the linear classifier in various combinations. We present our results in Table \ref{tab:opt}. The LARS optimizer \cite{you2017large} gives us the best results to train the embedding network, confirming what has been reported by previous work \cite{chen2020simple}. With LARS we use a cosine learning rate decay. On the other hand we find that the RMSProp optimizer \cite{tieleman2012lecture} works best for training the linear classifier. For RMSProp we use an exponential decay for the learning rate. 

\begin{table}[ht]
    \centering
    \begin{tabular}{ccc}\toprule
        {Contrastive Optimizer} & Linear Optimizer & Top-1 Accuracy \\\midrule
        LARS & LARS & 78.2 \\
        LARS & RMSProp & 78.7 \\
        LARS & Momentum & 77.6 \\ 
        RMSProp & LARS & 77.4 \\
        RMSProp & RMSProp & 77.8 \\
        RMSProp & Momentum & 76.9 \\
        Momentum & LARS & 77.7 \\
        Momentum & RMSProp & 76.1 \\
        Momentum & Momentum & 77.7 \\ \bottomrule
    \end{tabular}
    \vspace{2mm}
    \caption{Results of training the ResNet-50 architecture with AutoAugment data augmentation policy for 350 epochs and then training the linear classifier for another 350 epochs. Learning rates were optimized for every optimizer while all other hyper-parameters were kept the same.}
    \label{tab:opt}
\end{table}

\subsection{Data Augmentation}
We experiment with the following data augmentations:
\begin{itemize}
    \item {{\bf AutoAugment}: \cite{cubuk2019autoaugment} A two stage augmentation policy which is trained with reinforcement learning for Top-1 Accuracy on ImageNet.}
    \item {{\bf RandAugment}: \cite{cubuk2019randaugment} A two stage augmentation policy that uses a random parameter in place of parameters tuned by AutoAugment. This parameter needs to be tuned and hence reduces the search space, while giving better results than AutoAugment.}
    \item {{\bf SimAugment}: \cite{chen2020simple} An augmentation policy which applies random flips, rotations, color jitters followed by Gaussian blur. We also add an additional step where we  warping the image before the Gaussian blur, which gives a further boost in performance.}
    \item {{\bf Stacked RandAugment}: \cite{tian2020makes} An augmentation policy which is based on RandAugment \cite{cubuk2019randaugment} and SimAugment \cite{chen2020simple}. The strategy involves an additional RandAugment step before doing the color jitter as done in SimAugment. This leads to a more diverse set of images created by the augmentation and hence more robust training which generalizes better. }
\end{itemize}

and found that AutoAugment \cite{lim2019fast} gave us the highest Top-1 accuracy on ResNet-50 for both the cross entropy loss and supervised contrastive loss. On the other hand Stacked RandAugment \cite{tian2020makes} gives us highest Top-1 accuracy on ResNet-200 for both the cross entropy loss and supervised contrastive Loss. We conjecture this is happens because Stacked RandAugment is a stronger augmentation strategy and hence needs a larger model capacity to generalize well. 
 
We also note that AutoAugment is faster at runtime than other augmentation schemes such as RandAugment \cite{cubuk2019randaugment}, SimAugment \cite{chen2020simple} or StackedRandAugment \cite{tian2020makes} and hence models trained with AutoAugment take lesser time to train. We leave experimenting with MixUp \cite{zhang2017mixup} or CutMix \cite{yun2019cutmix} as future work.

\begin{table}[ht]
    \centering
    \begin{tabular}{ccc}\toprule
        Contrastive Augmentation & Linear classifier Augmentation & Accuracy  \\\midrule
        AutoAugment & AutoAugment & 78.6 \\
        AutoAugment & RandAugment & 78.1 \\
        AutoAugment & SimAugment & 75.4 \\
        AutoAugment & Stacked RandAugment & 77.4 \\\midrule
        SimAugment & AutoAugment & 76.1 \\ 
        SimAugment & RandAugment & 75.9 \\
        SimAugment & SimAugment & 77.9 \\
        SimAugment & Stacked RandAugment & 76.4 \\\midrule
        RandAugment & AutoAugment & 78.3 \\ 
        RandAugment & RandAugment & 78.4 \\ 
        RandAugment & SimAugment & 76.3 \\
        RandAugment & Stacked RandAugment & 75.8 \\\midrule
        Stacked RandAugment & AutoAugment & 78.1 \\
        Stacked RandAugment & RandAugment & 78.2  \\
        Stacked RandAugment & SimAugment & 77.9 \\
        Stacked RandAugment & Stacked RandAugment & 75.9 \\ \bottomrule
    \end{tabular}
    \vspace{2mm}
    \caption{Combinations of different data augmentations for ResNet-50 trained with optimal set of hyper-parameters and optimizers. We observe that stacked RandAugment does consistently worse for all configurations due to lower capacity of ResNet-50 models. We also observe that for other augmentations that we get the best performance by using the same augmentations in both stages of training. }
    \label{tab:aug}
\end{table}

\begin{figure}[h!]
    \centering
    \includegraphics[width=0.45\linewidth]{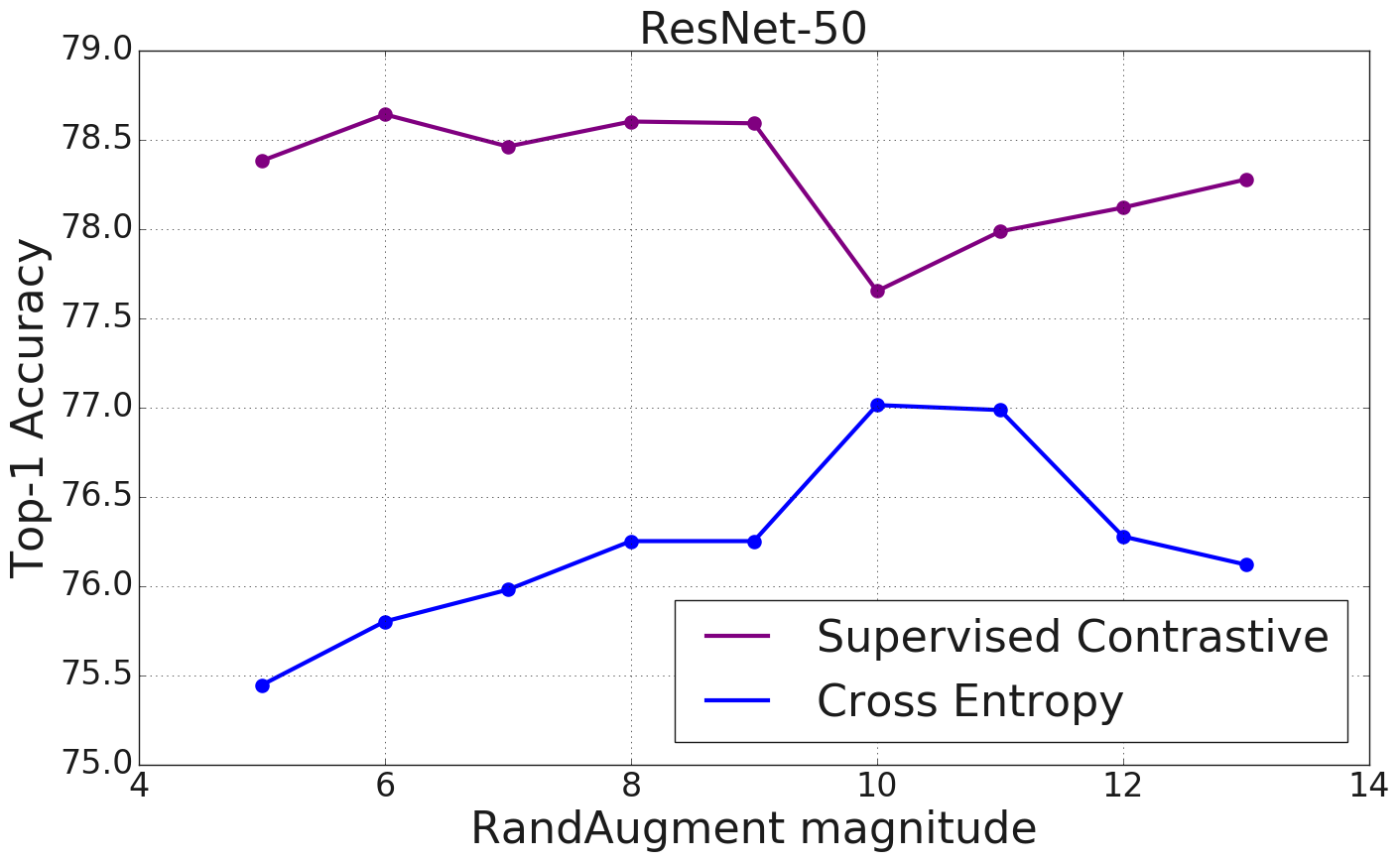}
    \includegraphics[width=0.45\linewidth]{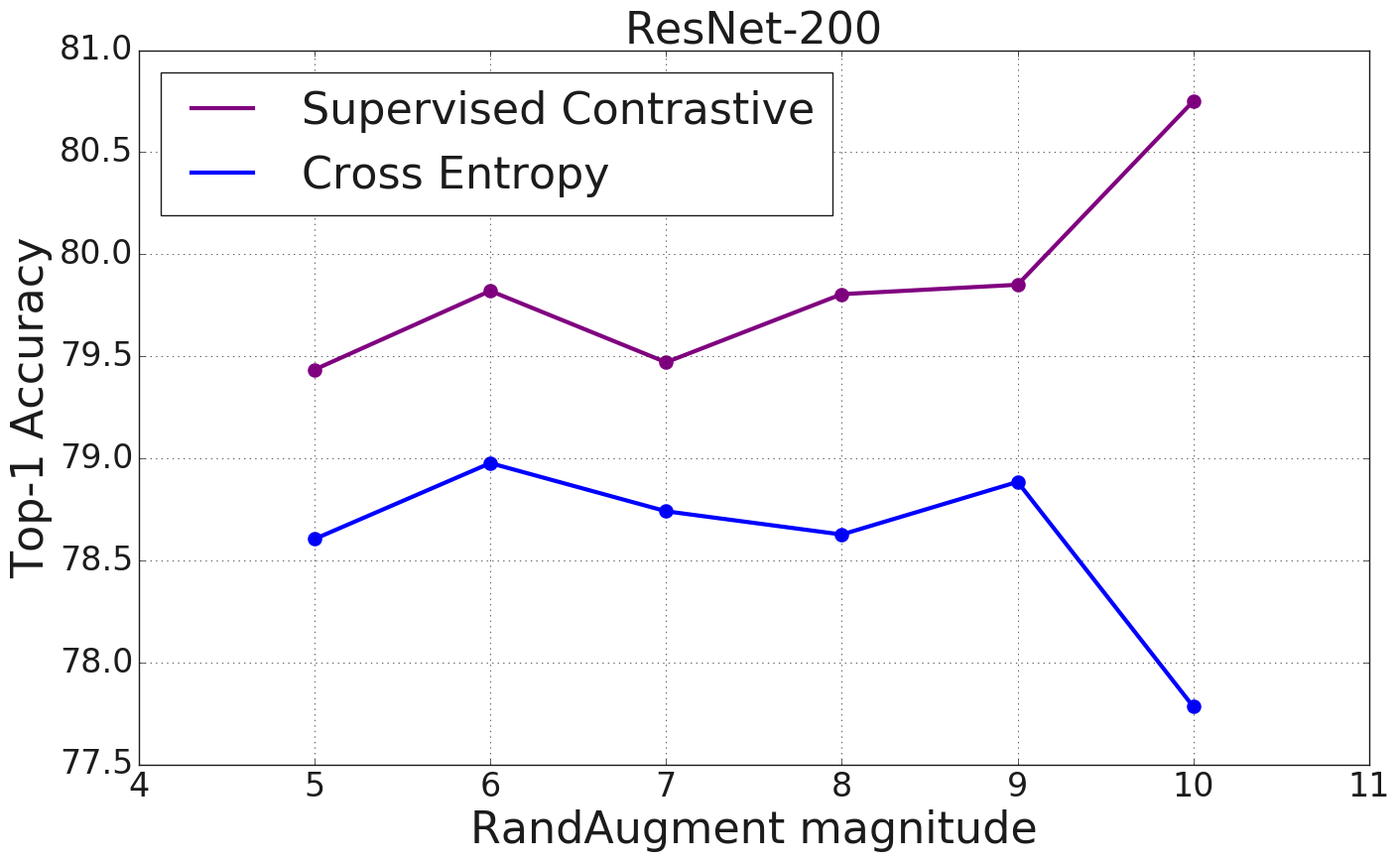}
    \caption{Top-1 Accuracy vs RandAugment magnitude for ResNet-50 (left) and ResNet-200 (right). We see that supervised contrastive methods consistently outperform cross entropy for varying strengths of augmentation.}
    \label{fig:randaug} 
\end{figure}

Further we experiment with varying levels of augmentation magnitude for RandAugment since that has shown to affect performance when training models with cross entropy loss \cite{cubuk2019randaugment}. Fig. \ref{fig:randaug} shows that supervised contrastive methods consistently outperform cross entropy training independent of augmentation magnitude.

\pagebreak
\section{Change Log}
\noindent\textbf{Version 1 (2020-04-23)}
Initial Arxiv version.

\vspace{3mm}
\noindent\textbf{Version 2 (2020-10-22)}
Added analysis of different forms of supervised contrastive loss and its gradients as well as experimental results to back up the analysis.
Additional tuning for better top-1 accuracy.
Added transfer learning experiments.
Moved accuracy vs num positives to supplementary.
More heavily tuned models resulted in deterioration of ECE.
Added StackedRandAugment augmentation.
Added GitHub link for code.
Added results vs batch size and number of training epochs.
Added results for more optimizers and data augmentation strategies.
Added SupCon loss hierarchy.
Adjusted table reporting for clarity.

\vspace{3mm}
\noindent\textbf{Version 3 (2020-10-13)}
Removed deprecated sentence from abstract.

\vspace{3mm}
\noindent\textbf{Version 4 (2020-12-10)}
Fixed a few in-paper equation references incorrectly pointing to equations in the Supplementary.

\vspace{3mm}
\noindent\textbf{Version 5 (2020-12-27)}
Added footnotes for authors on first page. 

\vspace{3mm}
\noindent\textbf{Version 6 (2020-03-10)}
Corrected sign mistake in Jensen's Inequality.

\end{document}